\documentclass[journal]{IEEEtran}

\usepackage[utf8]{inputenc} 
\usepackage[T1]{fontenc} 
\usepackage{color} 
\usepackage{url} 
\usepackage{glossaries} 
\usepackage{xspace} 
\usepackage{amsmath}  
\usepackage{amssymb}  
\usepackage{amsfonts} 
\usepackage{bm} 
\usepackage{nicefrac} 

\usepackage{graphicx} 
\usepackage{subcaption}

\usepackage{booktabs} 
\usepackage{tabulary} 
\usepackage{multirow} 

\newacronym{aa}{AA}{Adversarial Attack}
\newacronym{ai}{AI}{Artificial Intelligence}
\newacronym{auc}{AUC}{Area Under the Reciever Operating Curve}
\newacronym{asr}{ASR}{Attack Success Rate}

\newacronym{bim}{BIM}{Basic Iterative Method}

\newacronym{cam}{CAM}{Coverage Analysis Method}
\newacronym{cnn}{CNN}{Convolutional Neural Network}
\newacronym{cw}{CW}{Carlini \& Wagner}

\newacronym{df}{DF}{DeepFool}
\newacronym{dknn}{DkNN}{Deep k-Nearest Neighbors}
\newacronym{dnn}{DNN}{Deep Neural Network}
\newacronym{dmd}{DMD}{Deep Mahalanobis Detector}
\newacronym{dmd-a}{DMD-a}{DMD aware}
\newacronym{dmd-u}{DMD-u}{DMD unaware}
\newacronym{dmd-b}{DMD-b}{DMD base}
\newacronym{doc}{DOC}{Detecting Out-of-distribution and misclassifications by Confident score calibration}
\newacronym{dum}{DUM}{Deterministic Uncertainty Measurement}
\newacronym{dpa}{DPA}{Decision Process Analysis}

\newacronym{eu}{EU}{European Union}
\newacronym{ece}{ECE}{Expected Calibration Error}

\newacronym{fs}{FS}{Feature Squeezing}

\newacronym{gmm}{GMM}{Gaussian Mixture Model}
\newacronym{gp}{GP}{Gaussian Process}

\newacronym{hlf}{HLF}{High-Level Feature}

\newacronym{id}{ID}{In-Distribution}

\newacronym{km}{KM}{K-Means}
\newacronym{knn}{k-NN}{k Nearest Neighbors}
\newacronym{kl}{KL}{Kullback–Leibler}

\newacronym{llf}{LLF}{Low-Level Feature}
\newacronym{lrp}{LRP}{Layer-wise Relevance Propagation}
\newacronym{llm}{LLM}{Large Language Models}

\newacronym{ml}{ML}{Machine Learning}
\newacronym{msp}{MSP}{}

\newacronym{nlm}{NLM}{Non Local Mean Filter}

\newacronym{ood}{OOD}{Out-Of-Distribution}

\newacronym{pgd}{PGD}{Projected Gradient Descent}

\newacronym{relu}{Rel-U}{Relative Uncertainty}

\newacronym{svd}{SVD}{Singular Value Decomposition}
\newacronym{sota}{SOTA}{State-Of-The-Art}

\newacronym{fpr@95}{FPR@95}{False Positive Rate at 95\% True Positive Rate}

\newacronym{vit}{ViT}{Vision Transformers}
\newacronym{vgg}{VGG}{Visual Geometry Group}

\newacronym{xai}{XAI}{eXplainable Artificial Intelligence}

\newacronym{ourframework}{MACS}{Multi-layer Analysis for Confidence Scoring}

\newcommand{\circlemark}{\text{\Large$\circ$}}

\newcommand{\codeFont}[1]{\texttt{#1}\xspace}
\newcommand{\cifar}[1]{CIFAR-#1}

\newcommand{\places}{Places365\xspace}
\newcommand{\SVHN}{SVHN\xspace}
\newcommand{\vgg}[1]{VGG#1}
\newcommand{\vit}[2]{ViT#1#2}
\newcommand{\pytorch}{PyTorch}
\newcommand{\torchVision}{TorchVision}

\newcommand{\softmax}{softmax\xspace}

\newcommand{\realSet}{\mathbb{R}}
\newcommand{\probSet}{\mathbb{P}}

\newcommand{\mat}[1]{\ensuremath{\bm{{#1}}}}
\renewcommand{\vec}[1]{\ensuremath{\bm{{#1}}}}

\DeclareMathOperator*{\argmax}{\ensuremath{arg\,max}}

\newcommand{\threshold}{\ensuremath{\delta}}
\newcommand{\setFont}[1]{\ensuremath{\mathcal{#1}}}
\newcommand{\norm}[1]{\ensuremath{\left\lVert#1\right\rVert}}

\newcommand{\img}{\mat{X}}
\newcommand{\numLabels}{\ensuremath{L}}
\newcommand{\numLayers}{\ensuremath{M}}
\newcommand{\predictionProb}{\vec{z}}
\newcommand{\nn}{\ell}

\newcommand{\weights}{\mat{W}}
\newcommand{\biases}{\mat{b}}
\newcommand{\act}{\vec{x}}
\newcommand{\actIn}{\act}
\newcommand{\actInSize}{n}
\newcommand{\actOut}{\vec{y}}
\newcommand{\actOutSize}{m}

\newcommand{\dataset}{\setFont{D}}
\newcommand{\trainingSet}{\dataset^{\rm train}\xspace}
\newcommand{\validationSet}{\dataset^{\rm val}\xspace}
\newcommand{\testSet}{\dataset^{\rm test}\xspace}
\newcommand{\trainingSetSize}{|\trainingSet|}

\newcommand{\SVDU}{\mat{P}}
\newcommand{\SVDV}{\mat{Q}}
\newcommand{\SVDS}{\mat{\Sigma}}
\newcommand{\SVDVtrim}{\SVDV'}
\newcommand{\SVDStrim}{\SVDS'}

\newcommand{\iLabel}{\ensuremath{l}}
\newcommand{\iInstance}{\ensuremath{t}}
\newcommand{\iCluster}{\ensuremath{i}}

\newcommand{\corevecToCluster}{\varsigma}
\newcommand{\corevecToTag}{\tau}

\newcommand{\projectMatrix}{\mat{B}}
\newcommand{\affineTransform}{\mat{A}}
\newcommand{\approxAffineTransform}{\affineTransform'}

\newcommand{\corevector}{corevector\xspace} 
\newcommand{\corevectors}{corevectors\xspace} 
\newcommand{\coreVec}{\vec{v}}
\newcommand{\coreVecSize}{\kappa}

\newcommand{\means}{\vec{\mu}}
\newcommand{\numClusters}{\ensuremath{C}}
\newcommand{\covariance}{\mat{K}}
\newcommand{\clusterProb}{\ensuremath{\phi}}

\newcommand{\clusteringMembership}{\vec{m}}
\newcommand{\clusteringMembershipElement}{\ensuremath{m}}
\newcommand{\clusteringFn}{\ensuremath{\gamma}}

\newcommand{\empiricalPosterior}{\mat{U}}

\newcommand{\empiricalPosteriorElement}{U}
\newcommand{\empiricalPosteriorRowElement}{u}

\newcommand{\ph}{\vec{g}}

\newcommand{\concept}{\mat{G}}
\newcommand{\classificationmap}{\codeFont{classification-map}} 
\newcommand{\classificationmaps}{\codeFont{classification-maps}} 

\newcommand{\protoclass}{\codeFont{proto-map}} 
\newcommand{\protoclasses}{\codeFont{proto-maps}} 
\newcommand{\protoc}{\ensuremath{\mat{P}}}
\newcommand{\protoSet}{\setFont{P}}

\newcommand{\protoThreshold}{\ensuremath{\threshold}}
\newcommand{\protoScore}{\ensuremath{\confScore}\xspace}

\newcommand{\confScore}{\ensuremath{s}}

\newcommand{\fpr}{\ensuremath{\mathrm{FPR}^{\star}}\xspace}

\newcommand{\AABudget}{\epsilon\xspace}

\newcommand{\dmdPosDataset}{\dataset^{+}_{\rm{DMD}}}
\newcommand{\dmdNegDataset}{\dataset^{-}_{\rm{DMD}}}

\begin{document}
	\bstctlcite{IEEEexample:BSTcontrol}
    
	\title{
     Multi-Layer Confidence Scoring for Detection of Out-of-Distribution Samples, Adversarial Attacks, and In-Distribution Misclassifications
	}
	
	\author{
		\IEEEauthorblockN{
			Lorenzo Capelli\IEEEauthorrefmark{1}, 
			Leandro de Souza Rosa\IEEEauthorrefmark{1},
			Gianluca Setti\IEEEauthorrefmark{3}\IEEEauthorrefmark{2},
			Mauro Mangia\IEEEauthorrefmark{1}\IEEEauthorrefmark{2}, 
			Riccardo Rovatti\IEEEauthorrefmark{1}\IEEEauthorrefmark{2}}
		
		\IEEEauthorblockA{
			\IEEEauthorrefmark{1}DEI,
			\IEEEauthorrefmark{2}ARCES, University of Bologna, Italy,
			\IEEEauthorrefmark{3}CEMSE, KAUST, Saudi Arabia
		} 
		\IEEEauthorblockA{
			\{l.capelli,~leandro.desouzarosa,~mauro.mangia,~riccardo.rovatti\}@unibo.it,			gianluca.setti@kaust.edu.sa
		}
	}

	\maketitle
		
\begin{abstract}
	The recent explosive growth in \acrlongpl{dnn} applications raises concerns about the black-box usage of such models, with limited transparency and trustworthiness in high-stakes domains, which have been crystallized as regulatory requirements such as the \acrlong{eu} \acrlong{ai} Act. 
    While models with embedded confidence metrics have been proposed, such approaches cannot be applied to already existing models without retraining, limiting their broad application.
    On the other hand, post-hoc methods, which evaluate pre-trained models, focus on solving problems related to improving the confidence in the model's predictions, and detecting \acrlong{ood} or \acrlongpl{aa} samples as independent applications.
    To tackle the limited applicability of already existing methods, we introduce \gls{ourframework}\glsreset{ourframework}, a unified post-hoc framework that analyzes intermediate activations to produce \classificationmaps.
    From the \classificationmaps, we derive a score applicable for confidence estimation, detecting distributional shifts and adversarial attacks, unifying the three problems in a common framework, and achieving performances that surpass the state-of-the-art approaches in our experiments with the \vgg{16} and \vit{b}{16} models with a fraction of their computational overhead.
\end{abstract}
	
	\section{Introduction} \label{sec: introduction}

Over the past decade, \gls{ai} has attracted substantial attention, especially due to the performances reached by \glspl{dnn} across many domains. 
Typically, \glspl{dnn} are used as black-box systems, making it challenging to understand how specific inputs lead to particular outputs and impossible to trace the origin of an error.
Furthermore, the lack of understanding limits both critical trustworthiness and transparency in \glspl{dnn}, both important requirements for high-risk domains such as healthcare, finance, and autonomous systems.
These concepts have been recently crystallized as the \gls{eu}'s \gls{ai} Act \cite{eu_ai_act_2024}, which categorizes systems affecting critical domains as high-risk and mandates transparency, explainability, and human oversight.

To enhance trust in human-\gls{ai} interaction, there is a growing need for \glspl{dnn} certification tools for addressing issues such as confidence estimation, i.e., determining whether \gls{id} samples are correctly classified by the \gls{dnn} \cite{granese2021doctor,dadalto2023data}, and \gls{ood} and \gls{aa} detection \cite{lee2018simple}.
While these issues are often studied separately in the literature, classification reliability is compromised in all three cases, and their shared underlying nature suggests the need for a unified detection framework. 
However, current methods for confidence estimation often fail when applied to \gls{ood} or \gls{aa} scenarios, and similarly, techniques designed for \gls{ood} or \gls{aa} detection have never been explored for confidence estimation.

Methods to tackle these problems are traditionally divided into two families:
ante-hoc, which focus on developing new \gls{dnn} models emphasizing trustworthiness;
and post-hoc methods that extract information from existing models, being more broadly applicable, and hence the focus of this paper.

Moreover, recent research goes beyond analyzing only the model's outputs, focusing on complementing \glspl{dnn}' prediction with intermediate activations analysis to enhance trustworthiness \cite{papernot2018deep,lee2018simple}.
Nevertheless, such methods are often limited to small models due to the activations' high dimensionality.

This paper introduces \gls{ourframework}, a framework for analyzing the intermediate activations produced by a pre-trained \gls{dnn} using classical signal-processing techniques, including a dimensionality reduction step to ensure scalability, an unsupervised clustering to associate activation space-level features with human-interpretable ones. \gls{ourframework} returns a confidence metric that quantifies the reliability of a \gls{dnn}'s internal decision-making process. The proposed methodology imposes no limitations on either the number of layers processed in parallel or the types of layers it can handle.
\gls{ourframework}’s architecture aligns with the \gls{eu}'s \gls{ai} Act's requirements by providing an audit-friendly and unified multi-layer evaluation applicable to confidence estimation, \gls{ood}, and \gls{aa} detection, further enhancing deployability in real-world systems constrained by regulatory or safety concerns.

The rest of this paper is organized as follows:
Section \ref{sec: related works} presents the related work.
Section \ref{sec: proposed method} formally defines the proposed framework.
Section \ref{sec: experimental setup} describes the applications, models, and datasets used for testing.
Section \ref{sec: results} presents results when applying the proposed framework for confidence estimation, and \gls{ood} and \gls{aa} detection.
Finally, Section \ref{sec: conclusions} concludes the paper.

	\section{Related Works}\label{sec: related works}

As the \gls{eu}'s \gls{ai} Act's requirements reflect on improving \glspl{dnn}'  robustness and trustworthiness, a vast amount of recent research focuses on confidence estimation, and \gls{ood} and \gls{aa} detection.
Below, we position our work relative to \gls{sota} approaches across several relevant axes.

\subsection{Confidence Estimation and Trustworthiness}

As \glspl{dnn} typically provide predictions regardless of their input, a confidence estimate capturing the reliability of the prediction is a direct way to improve the model's trustworthiness.

Historically, softmax-based confidence measures \gls{msp} \cite{hendrycks2017baseline} have been adopted, in the case of neural classifiers, since its output intrinsically encodes the probabilities of a sample belonging to the available classes.
However, it has been shown that, for modern \glspl{dnn} such outputs are poorly calibrated, especially convolution-based ones \cite{minderer2021revisiting}, necessitating specific training procedures, ante-hoc methods, or the post-hoc computation of calibration functions \cite{guo2017calibration, perez2022beyond}.
Nevertheless, calibration-based approaches may create undesirable shifts in the distribution of correctly- and miss-classified samples \cite{singh2021dark}, prejudice already calibrated classifiers \cite{wang2021rethinking}, and degrade under data scarcity \cite{balanya2024adaptive}, defeating the purpose of providing a reliable confidence score and leading recent research to focus on ante-hoc methods to solve these issues \cite{wang2021rethinking,clarte2023expectation}.

Opposing calibration, recent research focuses on using uncertainty estimates as a proxy for a confidence score.
Despite the mathematical soundness and application success of Bayesian estimators \cite{williams1995gaussian}, their prohibitive computational costs have led to the development of dropout-based training methods, resulting in \glspl{dnn} with embedded uncertainty estimation \cite{gal2016dropout}, which is implicitly and explicitly regularized \cite{wei2020implicit}.
Other ante-hoc methods focus on appending a \gls{gp} step to capture the epistemic uncertainty of data \cite{liu2020simple}, without which the capability of predicting \gls{ood} samples is reduced \cite{postels2021practicality}, or a classifier \cite{fort2021exploring}, used to fine-tune the \gls{dnn} to \gls{ood} detection.

On the other hand, post-hoc epistemic uncertainty estimation methods are an alternative to capturing relationships between data, acting similarly to calibration, as they fit a confidence estimate function based on the model's predictions over a training set.
An early approach to use a decision risk as a confidence estimation \cite{geifman2017selective}, which has been recently extended by \gls{doc} \cite{granese2021doctor}, and \gls{relu} \cite{dadalto2023data} through considering the distributions of correctly- and miss-classified samples by the model, creating a confidence estimate based on the \gls{kl} divergence, or on a relative epistemic uncertainty, respectively.
\gls{doc} and \gls{relu} are considered the \gls{sota} on confidence estimation based on the model's output, outperforming calibration methods and other approaches in the same family.

Opposing uncertainty-based methods, \gls{ourframework} is a post-hoc method that approaches confidence estimation from the intermediate activation analysis side, capturing intra-data dependencies in a feature association step.

\subsection{Intermediate Activation Analysis}

To analyze the model's decision process, recent research focuses on evaluating the model's intermediate activations.
A first example is \gls{dknn} \cite{papernot2018deep}, which applies a \gls{knn} on the activations extracted from different layers, yielding clusters of samples that enable a confidence score estimation based on the number of neighbors agreeing with the prediction, applicable to both \gls{ood} and \gls{aa} detection.

Focusing on \glspl{cnn}, \gls{dmd} \cite{lee2018simple} proposes computing a confidence score based on perturbations of a dimensionality-reduced representation of the intermediate activations, and training a regressor to detect \gls{ood} or \gls{aa} samples.
Nevertheless, \gls{dmd}'s shortcomings are its restriction to convolutional layers and the necessity of \gls{ood}/\gls{aa} samples to fit its regressor.
Most recently, \cite{rossolini2022increasing} proposes an \gls{aa} detection method that estimates a confidence score by comparing intermediate activations against \gls{cam}-based signatures derived from typical activation values for samples in a reference dataset.
\cite{liu2020explaining} proposes clustering the intermediate activations and performing an association step, leveraging empirical relationships between \glspl{llf} within the \gls{dnn}'s activation and \glspl{hlf} perceived by humans.

We highlight that analyzing intermediate activations directly is not a scalable approach, as modern \glspl{dnn} easily reach dimensionalities over $1$M, being detrimental for distance-based methods such as clustering.
To address this limitation, \gls{ourframework} uses a \gls{svd} dimensionality reduction step, which we show applies to both linear and convolutional, allowing leveraging the association between \glspl{llf} and \glspl{hlf} from \cite{liu2020explaining}. 

\subsection{Adversarial Attack Detection}

As \gls{aa} are subtle modifications that lead the model to provide an incorrect prediction without altering the input's semantic meaning to the human user, they pose a reliability threat to systems deployed in practical and safety-critical applications.

In image classification tasks, \glspl{aa} are generated by applying small perturbations to the input image.
As such, approaches for \glspl{aa} detection often rely on filtering to generate multiple samples, performing detection based on the model's outputs to these samples.
One of the most well-established approaches in this family is \gls{fs} \cite{xu2017feature}, which applies multiple filters on the input and compares the model's decision on the computed variants against the original one to identify the presence of an \gls{aa}.
Similarly, \cite{liang2021detecting} applies a smoothing filter based on the sample's entropy, and differently, \cite{jha2019attribution} creates variants through perturbations.
The main limitations of these methods are that they are applicable only to image classifiers, and attacks can be devised to avoid detection \cite{he2017adversarial}.

Regarding intermediate activations analysis, early approaches on \glspl{aa} propose creating a complementary classifier fitted with normal and attacked samples on the intermediate activations \cite{metzen2017detecting}.
Furthermore, the previously mentioned \gls{dknn} \cite{papernot2018deep} and \gls{dmd} \cite{lee2018simple} also apply in this scenario.

\gls{ourframework} enables efficient adversarial detection by comparing the model's decision process against expected values for the attacked prediction.

\subsection{Reference Approaches}\label{subsec: reference approaches}

%


In this section, we position our method relative to the reference approaches.
Because it leverages intermediate activation analysis, \gls{ourframework} can address confidence estimation, \gls{ood} detection, and \gls{aa} detection within a unified scheme, offering greater flexibility than existing methods.

\subsubsection*{\gls{msp}}

As the model itself is its best estimator, its outputs are typically the most efficient metric for confidence estimation.
This makes \gls{msp} \cite{hendrycks2017baseline} a strong baseline for misclassifications detection, which is computed as the maximum value of the model's output after a softmax normalization.

\subsubsection*{\gls{doc}}

\gls{doc} \cite{granese2021doctor} is a post-hoc method for confidence estimation and \gls{ood} detection.
It computes a score based on the Gini's impurity of the model’s \softmax output and has two tunable hyper-parameters: temperature scaling and magnitude.
%
The former sharpens the model's output based on its distribution over a validation set.
When the magnitude is set to a positive value, the method applies a gradient-based input perturbation, similar to an adversarial processing step, highlighting features that influence the model’s confidence.

\subsubsection*{\acrlong{relu}}

\gls{relu} \cite{dadalto2023data} introduces a data-driven measure of relative uncertainty for confidence estimation and \gls{ood} detection.
It learns a class-wise distance matrix from correctly and incorrectly classified samples over a validation set, assigning higher uncertainty to misclassified instances.
The resulting score is computed in closed form from the \softmax outputs, after a temperature scaling step, which needs to be tuned. 
Similar to \gls{doc}, \gls{relu} employs an input pre-processing step controlled by a magnitude hyper-parameter, which back-propagates the learned metric to enhance separability.
%
Through a third hyperparameter, the proposed method balances the trade-off between minimizing uncertainty for correctly classified samples and maximizing it for misclassified ones.

\subsubsection*{\acrlong{dmd}}

\gls{dmd} \cite{lee2018simple} models the intermediate activations of \glspl{cnn} distribution for detecting \gls{ood} and \gls{aa} samples.
The authors introduce a simplified variant, which we referred to as \gls{dmd-b},  and which analyzes the statistical distribution of pre-logits activations.
As proposed in the previous approaches, back-propagation is applied to create a variation of the original sample based on the minimum Mahalanobis distance of the pre-logits activation with respect to each class training distribution. 
The framework is subsequently extended to a multi-layer formulation by applying the same analysis to convolutional layers after reducing them via average pooling.
%
The evaluation of all layers' scores is done empirically, by training a logistic regressor on the scores extracted from each layer to differentiate the vector of scores over a positive dataset ($\dmdPosDataset$), usually the \gls{id} validation set, and a negative one ($\dmdNegDataset$), typically \gls{ood} or \gls{aa} validation samples.

%
The multi-layer version can be set in two modes:
\gls{dmd-a} is set by using samples from the same dataset it is tested on as $\dmdNegDataset$, hence being a supervised method with expected high-performance, which we interpret as an upper-bound;
\gls{dmd-u} is set by using samples from a dataset different than the one it is tested on as $\dmdNegDataset$, hence being a realistic case in which the nature of $\dmdNegDataset$ cannot be defined a priori.

While the multi-layer analysis was originally designed for \glspl{cnn}, we extend it to transformer-based \glspl{dnn} by omitting the initial dimensionality-reduction step and focusing solely on the embedding activations that is associated with classification.

\subsubsection*{\acrlong{fs}}

\gls{fs} \cite{xu2017feature} assumes that the perturbation introduced by \glspl{aa} can be effectively removed by image pre-processing techniques, computing a score based on the distance between the model's output for the original image and its pre-processed versions.

We consider the three filters proposed on \cite{xu2017feature}, namely:
Bit Reduction, which uniformly quantizes values by discarding their least-significant bits;
Median Filter, a standard filter for noise removal based on the median operator;
and \gls{nlm}, a weighted denoising that averages pixels whose surrounding patches are similar. 
While the former is defined by the number of bits used to quantize the original input, the median filter depends on the filter's kernel size. 
\gls{nlm} pre-processing is determined by parameters such as the search window size, the patch size for similarity computation, and the Gaussian filter bandwidth controlling the smoothing.

We highlight that \gls{fs} is originally designed for \gls{aa} detection, and its efficiency depends on selecting filters tailored for the attacks to be detected.
Moreover, the approach is model-agnostic as the selected filters are created without using information about the \gls{dnn} under analysis.

	\section{Proposed Method}\label{sec: proposed method}

We deal with \glspl{dnn} classifier accepting input images encoded as tensors $\img$ and outputting one of $\numLabels$ possible labels $\nn(\img)=\iLabel$ with $\iLabel\in\{0,\dots,\numLabels-1\}$.
We assume that this happens through the computation of an intermediate vector $\predictionProb(\img)\in\probSet^\numLabels$ of positive numbers whose sum is normalized to $1$ that estimates the match between $\img$ and each possible label, so that $\nn(\img)=\argmax\predictionProb(\img)$.

The entries of $\predictionProb(\img)$ provide information about how sharp the network’s final decision is: a vector dominated by a single large entry, with all other components negligible, indicates that the network assigns overwhelming confidence to the chosen label; by contrast, a vector with several entries close to the maximum suggests that alternative labels are regarded as nearly as plausible as the output one. 

The model is composed of a sequence of intermediate layers with arbitrarily shaped inputs $\actIn$ and outputs $\actOut$, referred to as activations.
We assume the availability of datasets composed of input–label pairs, divided into {\em training} set $\trainingSet$, {\em validation} set $\validationSet$, generally used for tuning, and {\em test} set $\testSet$, used for performance evaluation, whose cardinalities are denoted by $|\cdot|$.
For instance, the number of input–label pairs in the training set is $\trainingSetSize$, and the set is represented as\footnote{The index $\iInstance$ is used to enumerate instances within any of the aforementioned sets. When omitted, it refers to a generic element of the set; when used without a specified range, it is assumed to span the entire set.} ${\{\img_\iInstance, \iLabel_\iInstance\}}_{\iInstance=0}^{\trainingSetSize-1}$.


\begin{figure*}
	\centering
	\includegraphics{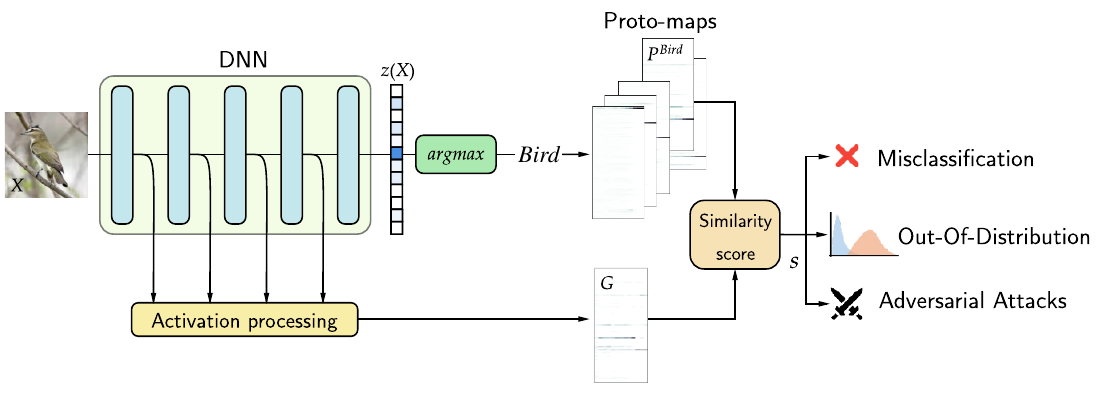}
	\caption{\gls{ourframework}'s overview. The intermediate activations are processed, leading to a compact representation of the model's decision process ($\concept$), which is compared with typical values for the predicted class ($\protoc$) resulting in a confidence score ($\protoScore$) that can be used for detecting misclassified samples, \gls{ood} inputs or \glspl{aa}.}
	\label{fig:generic_nn}
\end{figure*}

The main idea of the proposed framework is to compute a score that captures the coherence between patterns of intermediate activations and the network’s output, as visualized in Figure \ref{fig:generic_nn}.
The activation processing, described in detail in Section \ref{sec: peephole generation}, is performed on a set of target layers, and consists of:
\textit{i}) a dimensionality reduction step transforms high-dimensional activations into low-dimensional ones, namely \corevectors, capturing and compressing the layer's parameters for efficient downstream processing while minimizing information loss;
\textit{ii}) a clustering step identifies patterns within the \corevectors space;
\textit{iii}) an empirical association step relating the identified clusters with the classification labels, creating a vector that estimates the classification probabilities ($\predictionProb(\img)$) from the features encoded in a single layer.
By gathering these estimates from the set of target layers, we create the proposed \classificationmap ($\concept$), which maps the decision process throughout the \gls{dnn}'s layers and is compared against typical values for the predicted class ($\protoc$) to compute the proposed confidence score ($\protoScore$), as described in detail in Section \ref{subsec: multiple peepholes}.

\subsection{Internal Activation Processing} \label{sec: peephole generation}

As analyzing raw intermediate activations is computationally prohibitive, the first processing step is computing a low-dimensional vector that captures the information flowing through a layer.
We focus on layers accepting an input $\actIn \in \realSet^{\actInSize}$ and producing an output $\actOut \in \realSet^{\actOutSize}$ via the affine transformation:
\begin{equation}\label{eq: preactivations}
	\actOut=\left[
		\weights | \biases 
	\right]
	\left[ \begin{matrix} 
		\actIn \\
		1
	\end{matrix}\right]=
    \affineTransform
    \left[ \begin{matrix} 
		\actIn \\
		1
	\end{matrix}\right],
\end{equation}
where $\weights \in \realSet^{\actOutSize \times \actInSize}$ and $\biases \in \realSet^{\actOutSize}$ contain weights and biases, and $\affineTransform \in \realSet^{\actOutSize \times (\actInSize+1)}$ is the overall affine operator.
Note that eq. \eqref{eq: preactivations} is general and applies to layers expressible through parameter rearrangements, e.g., the isomorphic unrollings of convolutional layers \cite{praggastis2022svd} or \gls{vit} \cite{dosovitskiy2020image} ones.
Non-linear activation functions, normalization, or aggregation steps are considered as separate layers here.
%

A reduced representation of $\actIn$ is obtained by projecting it onto a $\coreVecSize$-dimensional subspace of $\realSet^{\actInSize+1}$ through a column-orthonormal matrix $\projectMatrix \in \realSet^{(\actInSize+1) \times \coreVecSize}$, i.e.,
$$
	\coreVec=\projectMatrix^\top
	\left[\begin{matrix}\act\\1\end{matrix}\right].
$$

The vector $\coreVec$, hereinafter referred to as \corevector, approximates the original output activations if an alternative transformation $\approxAffineTransform \in \realSet^{\actOutSize \times \coreVecSize}$ yields approximate outputs, i.e., minimizing
$$
	\left\|
		\approxAffineTransform\coreVec-\affineTransform
		\left[\begin{matrix}
			\act \\
			1
		\end{matrix}\right]
	\right\|^2 = 
	\left\|
		\approxAffineTransform\projectMatrix^\top
		\left[\begin{matrix}
			\act \\
			1
		\end{matrix}\right]
	-\affineTransform\left[\begin{matrix}\act\\1\end{matrix}\right]
	\right\|^2,
$$
where $\|\cdot\|$ is the classical square vector norm.

Averaging over random, zero-mean, and unit-variance inputs reduces this to minimizing $\|\approxAffineTransform \projectMatrix^\top - \affineTransform\|_F^2$, where $\|\cdot\|_F$ is the Frobenius norm.
The optimal solution is obtained from the \gls{svd} of $\affineTransform$:
$$
\affineTransform = \SVDU \SVDS \SVDV^\top,
$$
with $\SVDU$ and $\SVDV$ orthonormal and $\SVDS$ containing non-increasing singular values.
Let $\SVDVtrim\in\realSet^{(\actInSize+1)\times \coreVecSize}$ and $\SVDStrim\in\realSet^{\actOutSize\times \coreVecSize}$ denote the matrices formed by the first $\coreVecSize$ columns of $\SVDV$ and $\SVDS$, respectively.
From the \gls{svd} properties, $\projectMatrix = \SVDVtrim$ and $\approxAffineTransform = \SVDU \SVDStrim$, making the \corevectors to be computes as:
\begin{equation}\label{eq: corevector}
    \coreVec=\SVDVtrim^\top
	\left[\begin{matrix}
		\act\\1
	\end{matrix}\right]
\end{equation}

The second step aims to statistically capture patterns within the \corevectors' space through unsupervised clustering, allowing their distinction from unseen ones, e.g., ones generated by \gls{ood} or \gls{aa} samples. 

As best practice, all \corevectors are normalized according to mean and variance obtained from $\trainingSet$, after which they are clustered using a \gls{gmm} \cite{dempster1977maximum}, which defines a map from \corevectors into a membership vector $\clusteringMembership\in\probSet^\numClusters$ with components:
\begin{equation}\label{eq:d_clustering}
	\clusteringMembershipElement_\iCluster = \frac{\clusteringFn_\iCluster(\coreVec)}{\sum_{j=0}^{\numClusters-1}\clusteringFn_j(\coreVec)}
	~;~ i\in\{0,\dots,\numClusters-1\},
\end{equation}
where
$$
\clusteringFn_\iCluster(\coreVec) = \frac{\clusterProb_\iCluster}{\sqrt{(2\pi)^{\coreVecSize} \det(\covariance_\iCluster)}} e^{-\frac{1}{2} (\coreVec - \means_\iCluster)^\top \covariance_\iCluster^{-1} (\coreVec - \means_\iCluster)},
$$
with $\means_\iCluster\in\realSet^\coreVecSize$ being cluster center, $\covariance_\iCluster\in\realSet^{\coreVecSize\times \coreVecSize}$ the covariance matrix, and $\clusterProb_\iCluster$ is the membership probability.

Although we focus on \gls{gmm}, we highlight that any clustering algorithm can be adopted, providing the definition of membership functions $\clusteringFn_i:\realSet^\coreVecSize\mapsto\realSet^+$ that yield increasing values the more likely $\coreVec$ is to belong to the $i$-th cluster.

The third step consists of associating clustering assignments, i.e., \gls{llf} in the compressed activation space, with human-understandable \gls{hlf} represented by the labels of each instance. 
As in \cite{liu2020explaining}, this relationship is modeled assuming two functions, $\corevecToCluster:\realSet^\coreVecSize\mapsto\{0,\dots,\numClusters-1\}$ and $\corevecToTag:\realSet^\coreVecSize\mapsto\{0,\dots,\numLabels-1\}$, that associate \corevectors to clusters and labels, respectively, with a probabilistic link between them expressed as:
\begin{equation}\label{eq: probabaility of a class}
	\Pr\left\{\corevecToTag(\coreVec)=\iLabel\right\} = 
	\displaystyle\sum_{i=0}^{\numClusters-1}
	\Pr\left\{\corevecToTag(\coreVec)=\iLabel|\corevecToCluster(\coreVec)=\iCluster\right\}
	\Pr\left\{\corevecToCluster(\coreVec)=i\right\},
\end{equation}
with $\iLabel=0,\dots,\numLabels-1$.
In eq. \eqref{eq: probabaility of a class} we may interpret the membership vector $\clusteringMembership$ in terms of probabilities to set $\Pr\left\{\corevecToCluster(\coreVec)=\iCluster\right\}=\clusteringMembershipElement_\iCluster$, while the matrix $\empiricalPosterior_{\iLabel,\iCluster}=\Pr\left\{\corevecToTag(\coreVec)=\iLabel|\corevecToCluster(\coreVec)=\iCluster\right\}$ needs to be estimated from data.

By considering inputs, labels and associated \corevectors as $\{\img_\iInstance,\iLabel_\iInstance,\coreVec_\iInstance\}_{\iInstance=0}^{\trainingSetSize-1}$, the number joint events $\empiricalPosteriorRowElement_{\iLabel,\iCluster}$ can be computed as:
$$
    \empiricalPosteriorRowElement_{\iLabel,\iCluster} = \left|\left\{\corevecToTag(\coreVec_t)=l \wedge \corevecToCluster(\coreVec_t)=i\right\}\right|,
$$
and each element of $\empiricalPosterior$, $\empiricalPosteriorElement_{\iLabel,\iCluster}$, can be estimated as:
$$
\empiricalPosteriorElement_{\iLabel,\iCluster}=\frac{\empiricalPosteriorRowElement_{\iLabel,\iCluster}}{\displaystyle\sum_{j=0}^{\numLabels-1}\empiricalPosteriorRowElement_{j,\iCluster}}.
$$

Once estimated, $\empiricalPosterior$ can be used to compute the vector $\ph=\empiricalPosterior\clusteringMembership$ which acts as a network output estimation.
In other words, $\ph\in\probSet^{\numLabels}$ estimates the probability of a sample belonging to each class (label) computed from the \gls{llf} within the respective layer's activations.

Note that the \corevectors dimension $\coreVecSize$ and number of clusters $\numClusters$ are the only two hyper-parameters controlling the $\ph$ extraction, influencing the clustering and, consequently, the capability of $\ph$ to act as an estimator of $\nn({\img})$.

\subsection{Classification Maps and Scoring Function}\label{subsec: multiple peepholes}

As illustrated in Figure \ref{fig:generic_nn}, the three-stage processing chain is attached to a selected set of target affine layers.
The extracted vectors $\ph_j$ are arranged as column-wise into the matrix $\concept = [\ph_j]_{j=0}^{\numLayers-1} \in \realSet^{\numLabels \times \numLayers}$, referred to as the \classificationmap, where $\numLayers$ denotes the number of layers from which a $\ph_j$ is extracted.
Note that it may be desirable to process activations of every affine layer; however, a trade-off between computational cost and retrieved information relevance suggests focusing on a subset of layers.
The \classificationmaps can be visualized as an image representing the internal signature of the network’s decision process.

In the final step, we create a scoring function that estimates the model’s confidence during inference by quantifying how well $\concept$ matches with reference \classificationmaps from training samples, hereafter denoted as \protoclasses, hence capturing typical decision processes well encoded within the model. 

Consider the \classificationmaps $\{\concept_\iInstance\}_{\iInstance=0}^{\trainingSetSize-1}$.
The \protoclass $\protoc^\iLabel$ for each class $\iLabel\in\{0,\dots,\numLabels-1\}$ accumulates information from \classificationmaps of correctly classified samples with high-confidence, i.e., samples well fitted withing the model, formally described by the set $\protoSet^l=\{\concept_\iInstance ~|~ \nn(\img_\iInstance)=\iLabel_t \land \max(\predictionProb(\img_\iInstance))>\protoThreshold\}$, with $\protoThreshold$ being an user-defined threshold for the model's confidence.

The \protoclass for each possible label $\protoc^\iLabel$ is computed by the accumulation:
$$
\overline{\protoc}^\iLabel = \sum_{\iInstance\in\protoSet^l}\concept_\iInstance,
$$
followed by a normalization across the labels direction, such that each column of $\protoc^\iLabel$ is an element of $\probSet^\numLabels$.

Note that the proposed \protoclasses are similar to the ``signature values'' in the \gls{cam} analysis done in \cite{rossolini2022increasing} w.r.t. the set conditions in $\protoSet^l$, differing as the former encodes typical decision process values for each label, while the latter directly encodes the high-dimensional intermediate activations.

Finally, the score of a sample is computed as the cosine similarity between its \classificationmap and the \protoclass of its predicted label, as:
\begin{equation}\label{eq: proto score}
	\protoScore_t = \frac{\protoc^{\nn(\img_t)}\odot\concept_t}{\norm{\protoc^{\nn(\img_t)}}_F \norm{\concept_t}_F},
\end{equation}
where $\odot$ is the element-wise product.
Note that $\protoScore\in[0,1]$ since $\protoc^{\nn(\img_t)}$ and $\concept_t$ contain only non-negative values.

As a final remark, we highlight that the \gls{svd} decomposition to compute $\SVDVtrim$, \corevectors clustering, $\empiricalPosterior$ matrix, and \protoclasses $\protoc$ are computed offline, only once over $\trainingSet$, making the computation of $\protoScore$ at inference time a relatively lightweight process.
	\section{Experimental Setup}\label{sec: experimental setup}

This section briefly describes the experimental setup used in this paper\footnote{Source code available at: \url{https://github.com/SSIGPRO/Peepholes-Analysis}}.

Our experiments consider the well-known \cifar{100} \cite{krizhevsky2009learning} dataset, containing images of $\numLabels=100$ fine-grained classes.
To evaluate performance under distributional shifts, we use \cifar{100C} \cite{hendrycks2019robustness}, which is obtained from \cifar{100}'s $\testSet$ by applying $13$ corruptions with $5$ intensity levels.
For \gls{ood} detection, we consider \places \cite{zhou2017places} and \SVHN \cite{netzer2011reading}, standard benchmarks semantically distant from \cifar{100}'s distribution \cite{yang2022fsood}.
%

The driving examples are the classical image classifiers from the \gls{vgg} \cite{simonyan2014very} and \gls{vit} \cite{dosovitskiy2020image} families.
Specifically, we use \vgg{16}, a $16$-layer \gls{cnn} and \vit{B}{16}, a $12$-transformer layer with $16\times16$ patches.
\vgg{16} is characterized by its sequential arrangement of convolutional layers, followed by fully connected ones for the final classification. 
\vit{B}{16} is an encoder-only transformer architecture characterized by $12$ encoders, each composed of a multi-head attention layer followed by a sequence of two fully connected layers.
Both models are implemented within the \torchVision \cite{marcel2010torchvision} library from \pytorch \cite{ansel2024pytorch}. Since the models are pre-trained on the ImageNet classification task, we replace the final fully connected layer with a new layer containing $\numLabels$ output units. The resulting architectures are then fine-tuned on the \cifar{100} $\trainingSet$.

Classification performance is quantified by the top-$1$ accuracy, i.e., the ratio between the number correctly classified and the total samples.
\vgg{16} and \vit{B}{16} reach accuracies of $77.00\%$ and $86.64\%$, respectively, on \cifar{100}'s $\testSet$.
For \vgg{16} we extended the internal activation analysis to $13$ out $16$ layers composing the neural network, excluding only the initial three convolutional layers. For \vit{B}{16}, we select both fully connected layers immediately after the attention mechanism of each encoder block and the final classification head, resulting in a total of 25 analyzed layers.

For each of these target layers, $\coreVecSize$ and $\numClusters$ are tuned to maximize the top-3 accuracy computed by comparing the position of the $3$ highest entries in $\ph_t$ and the true label $\iLabel_t$.
%

%


%
%
%

\subsection{Adversarial Attacks}\label{subsec: attacks}

To evaluate \gls{aa} detection, we select four established attack algorithms in literature: \gls{bim} \cite{BIM}, \gls{pgd} \cite{PGD}, \gls{cw} \cite{CW}, and \gls{df} \cite{DeepFool}. 
We use \gls{cw}'s original implementation provided by the authors \cite{CW} and the TorchAttacks library \cite{kim2020torchattacks} implementations for the other \glspl{aa}.
All attacks are untargeted, and the attack budget, $\AABudget$, is set at $\nicefrac{8}{255}$. The generation of adversarial samples relies on different distance metrics: the Euclidean norm is used for \gls{cw} and \gls{df}, whereas \gls{pgd} and \gls{bim} minimize the Chebyshev norm. Table \ref{tab: asr} reports the \gls{asr} for each model and attack under test, which corresponds to the ratio between the amount of adversarial samples generated from correctly classified samples that successfully mislead the \gls{dnn} and the total amount of correctly classified samples.

\begin{table}[h]
	\centering
	\caption{\gls{asr} ($\uparrow$) for the proposed attacks over \vgg{16} and \vit{B}{16}.}
	\label{tab: asr}
	\begin{tabulary}{\linewidth}{lcc}
		\toprule
		\textbf{\gls{aa}} & \textbf{\vgg{16}} & \textbf{\vit{B}{16}} \\
		\midrule
		
		\gls{bim}	& 0.84 &  0.68\\
		\gls{cw}  	& 1.00 &  1.00\\
		\gls{df}	& 0.82 &  0.53\\
		\gls{pgd}   & 0.95 &  0.89\\
		\bottomrule
	\end{tabulary}
\end{table}
    \section{Numerical Evidence}\label{sec: results}

This section evaluates \gls{ourframework} regarding its ability to improve the confidence of classifications, and to detect \gls{ood} and \gls{aa} images by comparing it against the \gls{sota} baselines: \gls{msp} \cite{hendrycks2017baseline}, \gls{doc} \cite{granese2021doctor}, \gls{relu} \cite{dadalto2023data}, \gls{fs} \cite{xu2017feature}, and \gls{dmd} \cite{lee2018simple}  in its three variants: \gls{dmd-b}, \gls{dmd-a}, and \gls{dmd-u}.
To contextualize this comparison, Table \ref{tab:SOTA_vs_US} summarizes the reference approaches and the applications they were designed to address.

As discussed in Section \ref{subsec: reference approaches}, \gls{doc} and \gls{relu} are not designed for \gls{aa} detection, nor \gls{fs} and \gls{dmd} for confidence estimation; however, these approaches are applicable and tested for these applications, highlighting their shortcomings and \gls{ourframework}'s flexibility.

Note that all approaches ultimately compute a confidence score, which we normalize between $[0,1]$, with $1$ meaning that the prediction should be trusted and $0$ that it should not be trusted.
As such, a low score suggests that the sample should not be trusted, i.e., it is \gls{ood}, generated by an \gls{aa}, or an incorrectly classified in-distribution sample.

To quantitatively assess the performance of the considered methods in the various scenarios, we compute \gls{auc} \cite{Fawcett_PATREC2006} and \fpr.
The former measures overall discrimination capability by evaluating performance across all possible confidence-score thresholds. 
In contrast, \fpr is obtained by selecting a threshold such that $95\%$ of correctly classified in-distribution samples have their scores above it, this reflects a practical acceptance/rejection criterion when deciding whether to trust a prediction.

\begin{table}[!htbp]
	\centering
	\caption{
		Proposed and \gls{sota} approaches applications. $\checkmark$ indicates that the method is designed for the task.
		$\circlemark$ indicates that the method was not designed for the task, but was included for comparison.
	}
	\label{tab:SOTA_vs_US}
	\begin{tabulary}{\linewidth}{L C C C}
		\toprule
		\textbf{Method} &  \textbf{Confidence} & \textbf{OOD} & \textbf{Attack Detection} \\ 
		\midrule
		\textbf{\gls{ourframework} (our)} & \checkmark& \checkmark & \checkmark \\
		\gls{relu} \cite{dadalto2023data} & \checkmark & \checkmark & \circlemark \\
        \gls{doc} \cite{granese2021doctor} & \checkmark & \checkmark & \circlemark \\
		\gls{dmd} \cite{lee2018simple} & \circlemark & \checkmark & \checkmark \\
		\gls{fs} \cite{xu2017feature} & \circlemark & \circlemark & \checkmark \\
		\bottomrule
	\end{tabulary}
\end{table}


\subsection{\protoclasses}\label{subsec: conceptograms}

Computing \protoScore as in \eqref{eq: proto score}  by matching the \protoclass $\protoc$ associated to the \gls{dnn} prediction with the \classificationmap $\concept$ obtained at inference time yields two principal advantages.
First, it enables a more refined assessment of the validity of the predicted label by leveraging internal network activations, without incurring the additional computational cost associated with back-propagation-based procedures, as in \gls{dmd}. 
Second, \protoScore captures the statistical structure of the decision-making process across the entire network, rather than relying on a single layer or solely on the softmax output.
Consequently, the proposed framework yields an unsupervised reliability estimate of the \gls{dnn}'s predictions, thereby obviating the need for example \gls{ood} or \gls{aa} samples for integrating multi-layer information, as required by \gls{dmd-a} or \gls{dmd-u}.

\begin{figure}
	\centering
	\includegraphics[width=\linewidth]{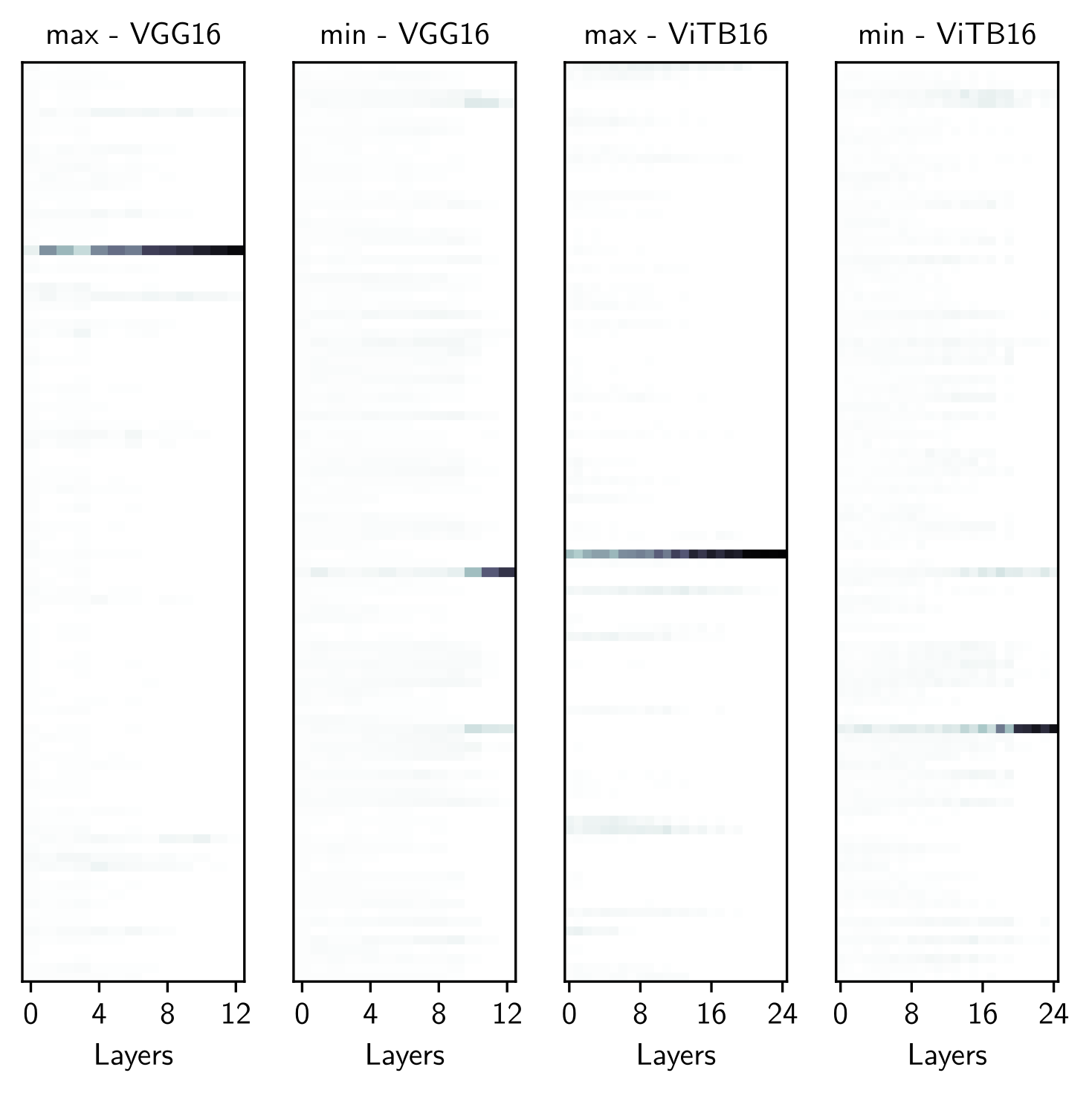}
	\caption{Examples of $\protoc$ computed from \vgg{16} and \vit{B}{16}. For each network, we display the proto-maps with the largest and smallest Frobenius norm. Columns correspond to the indices of the analyzed layers, while rows represent the dataset classes.}
  \label{fig: proto-maps}
\end{figure}

Figure \ref{fig: proto-maps} illustrates examples of $\protoc$ computed from \vgg{16} and \vit{B}{16}.
For each \gls{dnn} we selected $\protoc$ that have the maximum and minimum Frobenius norm. Each $\protoc$ captures the characteristic decision-making trajectory of the reference \gls{dnn} for correctly classified samples with high confidence belonging to a specific class. 
Note that each class induces a characteristic activation pattern, and the proposed score quantifies how much the $\concept$ deviates from that class-specific pattern. Importantly, its performance is not biased by the prevalence of the predicted class across the layers of $\protoc$.

\subsection{Scoring in-Distribution Inputs}\label{subsec: confidence}

In this initial evaluation, the confidence scores produced by all methods listed in Table \ref{tab:SOTA_vs_US} are employed to determine, in a straightforward manner, whether predictions are correct or incorrect.

Note that, in this application, \gls{dmd-a}'s regressor is trained on all wrongly classified samples, i.e., $\dmdNegDataset=\{\img_\iInstance  \in \trainingSet~|~\nn(\img_\iInstance)\neq\iLabel_\iInstance\}$, and an equally populated subset of the correctly classified samples, $\dmdPosDataset=\{\img_\iInstance \in \rm{shuffle}(\trainingSet)~|~\nn(\img_\iInstance)=\iLabel_\iInstance\}_{t=0}^{|\dmdNegDataset|}$. 

\begin{figure}
	\centering
	\includegraphics[width=\linewidth]{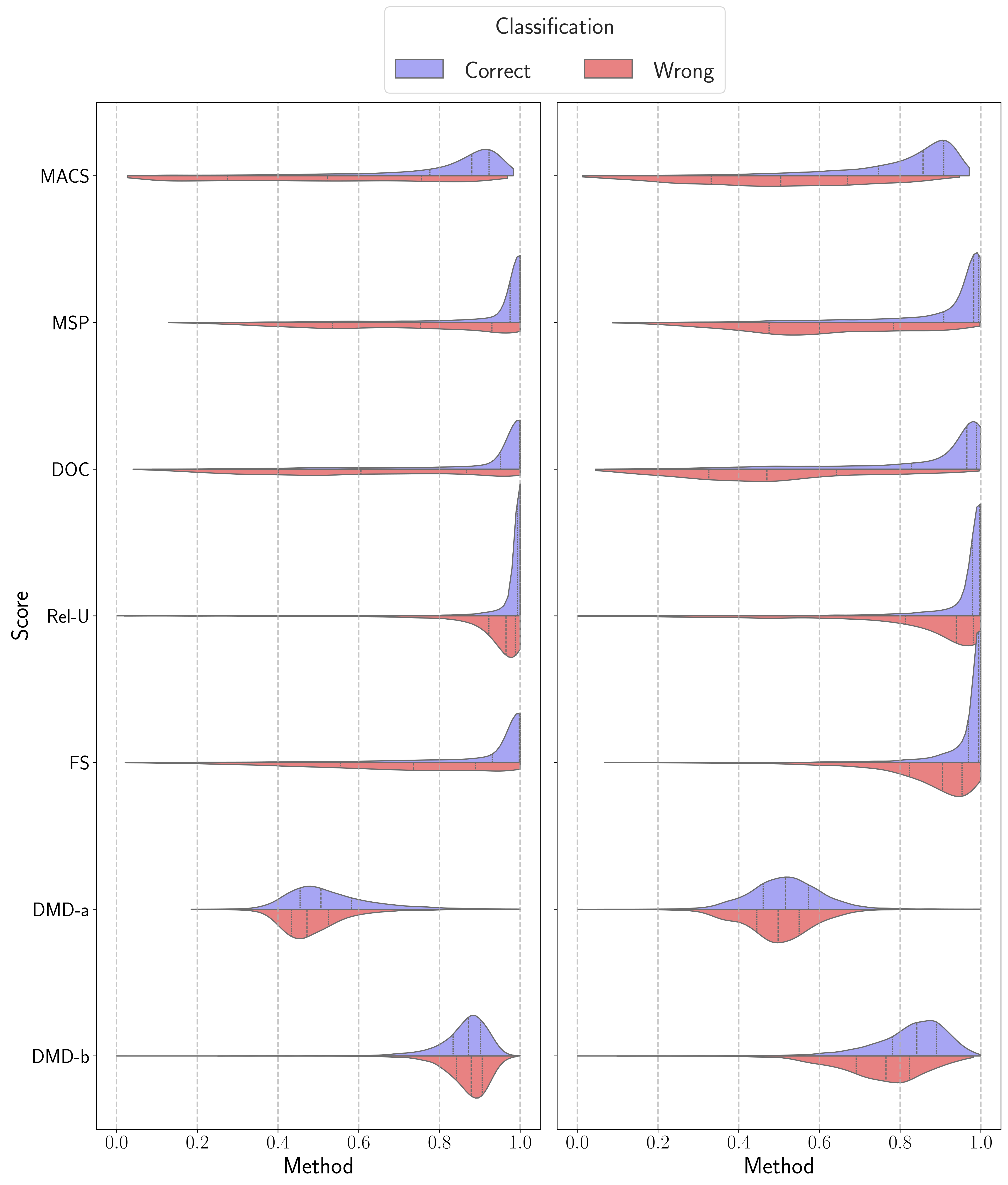}
	\caption{Confidence scores distribution for correctly and misclassified samples over \vgg{16} (left) and \vit{B}{16} (right).}
	\label{fig: confidence}
\end{figure}

Figure \ref{fig: confidence} shows the score distributions conditioned by correct and incorrect classifications\footnote{KDE approximation of the distributions.}. 
The profiles related to \gls{ourframework} evidence how the proposed score in \eqref{eq: proto score} exhibits less overconfidence than the other metrics, as $\protoScore$ rarely approaches a value of 1.
This property allows incorrectly classified samples to be pushed toward low-confidence regions of the distribution.
However, this advantage comes at the cost of assigning lower confidence scores to a portion of correctly classified samples, compared with the alternative methods.


To quantitatively evaluate the distributions in Figure \ref{fig: confidence}, Table \ref{tab: confidence metrics} reports the \glspl{auc} and {\fpr}s, showing that \gls{msp} achieves the best \glspl{auc}.
These results confirm that the model itself is a very good estimator, whereas the other methods provide only surrogate approximations of its predictive certainty.
Interestingly, among the methods not originally designed for this task, \gls{dmd} performs poorly in both single-layer \gls{dmd-b} and multi-layer \gls{dmd-a} versions, and \gls{fs} reaches good \glspl{auc}, but its high {\fpr}s indicate the shortcomings in generalizing the method to this task.
Regarding the other methods, \gls{doc} achieves the highest \glspl{auc}, and \gls{ourframework} the lowest \fpr in both cases, which is favorable in a practical scenario since it is more important to attribute a lower score for low-confidence samples than a high score for high-confidence ones.

\begin{table}
  \centering
  \caption{\gls{auc} ($\uparrow$) and \fpr ($\downarrow$) values for the proposed and reference methods over \vgg{16} and \vit{B}{16}.}
  \begin{tabular}{lcccc}
    \toprule
    \multirow{3}{*}{\textbf{Method}} & \multicolumn{2}{c}{\textbf{\vgg{16}}} & \multicolumn{2}{c}{\textbf{\vit{B}{16}}} \\ \cmidrule(lr){2-3} \cmidrule(lr){4-5}
    
      & \fpr &  \gls{auc} & \fpr & \gls{auc}  \\\midrule
      
    \gls{ourframework}	& 0.63 & 0.83 & 0.55 & 0.87  \\
    \gls{msp}   		& 0.65 & 0.87 & 0.57 & 0.90  \\
    \gls{doc}       	& 0.64 & 0.87 & 0.58 & 0.90  \\
    \gls{relu}        	& 0.80 & 0.85 & 0.78 & 0.82  \\
    \gls{fs}			& 0.76 & 0.85 & 0.75 & 0.86  \\
    \gls{dmd-b}         & 0.97 & 0.46 & 0.83 & 0.72  \\
    \gls{dmd-a}         & 0.92 & 0.61 & 0.90 & 0.56  \\
    \bottomrule
  \end{tabular}
  \label{tab: confidence metrics}
\end{table}

\begin{figure}
	\centering
	\includegraphics[width=\linewidth]{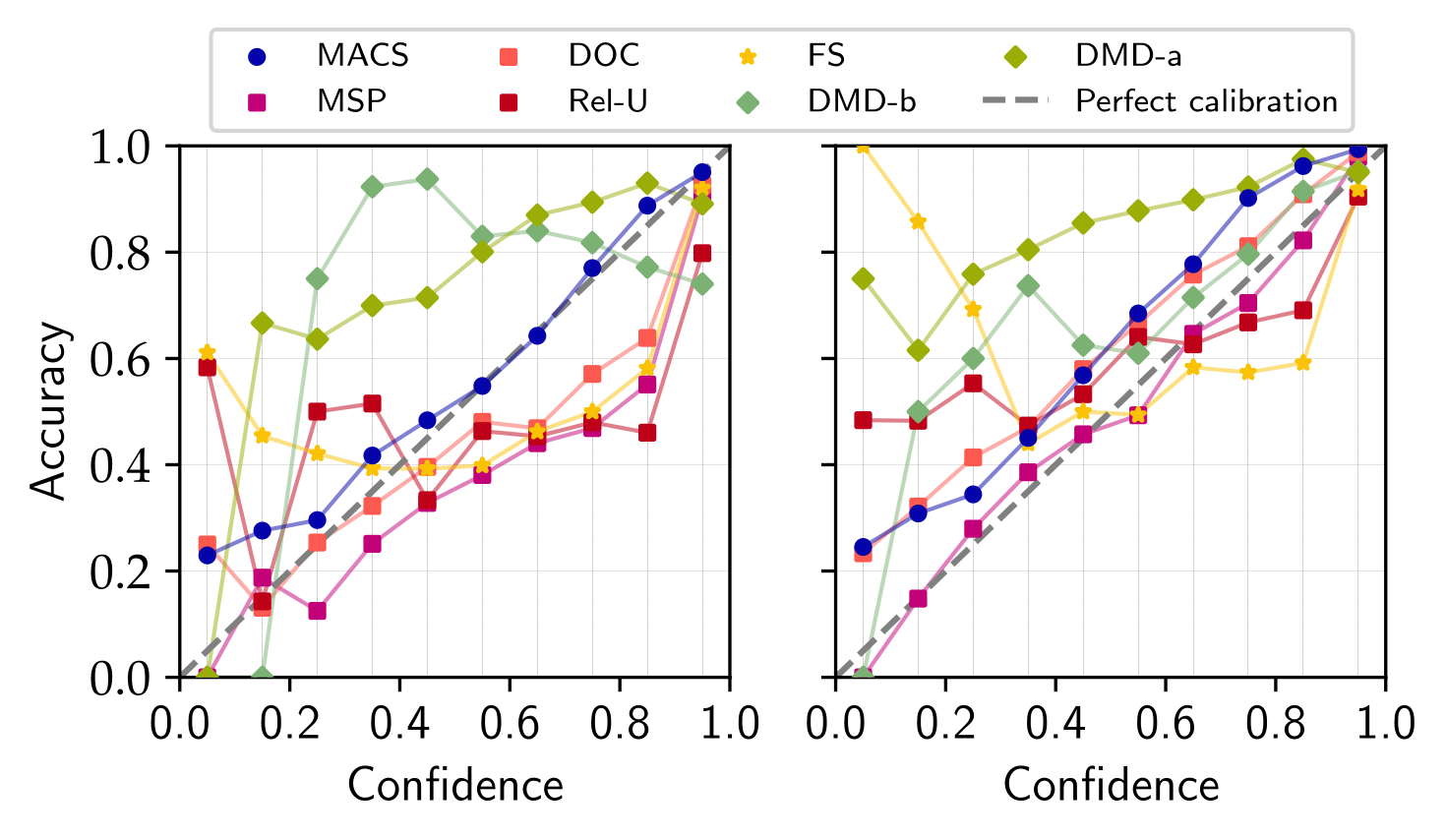}
	\caption{Reliability diagrams for \vgg{16} (left) and \vit{B}{16} (right).}
	\label{fig: calibration}
\end{figure}

The results in Figure \ref{fig: confidence} are tied to the scores' calibration, which evaluates the average accuracy of samples as a function of their scores, typically visualized as a reliability diagram, that bins samples by score and measures their average accuracy.
A perfectly calibrated score is a diagonal line, and a score below or above this line is over- or under-confident, respectively. 
The corresponding diagrams for \cifar{100} are shown in Figure \ref{fig: calibration}. 
Consistent with the conditional distributions in Figure \ref{fig: confidence}, the scores produced by \gls{ourframework} exhibit under-confidence.

%
Interestingly, \gls{ourframework} has accuracy that monotonically increases with the confidence for both cases, while the reference methods have undesirable jumps caused by the accuracy's high-variability under the limited number of samples with low confidence, ultimately demonstrating that the \gls{doc}, \gls{relu}, and \gls{fs} are not well calibrated.
In fact, even \gls{msp} shows these problems for \vit{B}{16}, even though a good calibration for this model was expected according to \cite{minderer2021revisiting}.
%

		

    
	

\subsection{Out-Of-Distribution Detection}\label{subsec: ood}

This section evaluates how well the scores allow for differentiating \gls{id} samples, i.e., typical images seen by the model during its training, from \gls{ood} ones, i.e., substantially different images which are not well encoded within the \gls{dnn}'s parameters. 

First, we consider the detection of corrupted samples using the \cifar{100C} dataset.
Note that \gls{dmd} is hard-limited to its aware mode (\gls{dmd-a}), meaning that corrupted samples (from \cifar{100C}'s $\validationSet$) are used as supervision to fit its regressor, hence being interpreted here as an upper-bound for unsupervised methods.



\begin{figure}
	\centering
	\includegraphics[width=\linewidth]{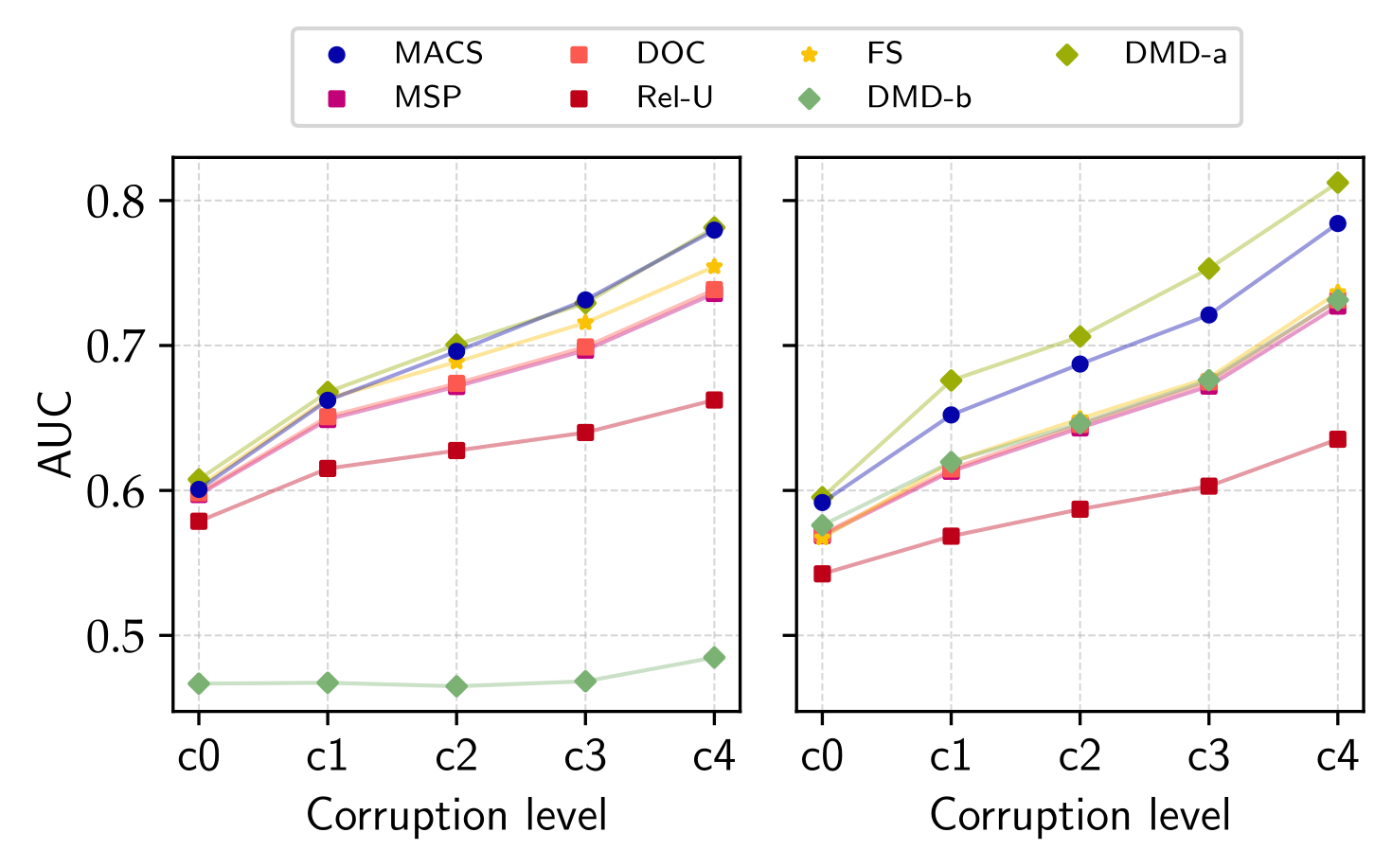}
    \caption{
        \gls{auc} progression with the corruption intensity using \gls{id} samples from \cifar{100}'s $\testSet$ and corrupted samples from \cifar{100C}, with $cX$ indicating the corruption intensity, for \vgg{16} (left) and \vit{B}{16} (right). As the intensity increases, scores decrease, making it easier to differentiate normal from corrupted samples.
    	}
    \label{fig: corruption}
\end{figure}

Figure \ref{fig: corruption} shows the \gls{auc} for each corruption intensity for samples from \cifar{100} and \cifar{100C}'s $\testSet$.
Note that \gls{msp} is over-confident for both types of samples, meaning that a high score is attributed to either case, regardless of the corruption intensity, making it a non-robust score for this task, as demonstrated by the results from \cite{liang2017enhancing}.
\gls{dmd-a} consistently yields good \glspl{auc}, which is expected since it is the only supervised method.
Regarding \gls{ourframework}, results show that it outperforms all other non-supervised methods for \vit{B}{16}, while reaching the same level of performance as \gls{dmd-a} on \vgg{16}.
It is worth noting that \gls{fs}, a method not designed for this task, outperforms \gls{doc} and \gls{relu}, even though its performance is below the proposed method on the modern model \vit{B}{16}.

Next, we evaluate an \gls{ood} scenario using \SVHN and \places.
In this case, \gls{dmd-u} is set by training the regressor with samples from the other dataset, i.e., $\dmdNegDataset=\validationSet_{\rm{\places}}$ for testing \SVHN and $\dmdNegDataset=\validationSet_{\rm{\SVHN}}$ for testing \places.
In turn, \gls{dmd-a} is set by fitting its regressor on $\dmdNegDataset$ using samples from the same dataset as it is tested on; i.e., $\dmdNegDataset=\validationSet_{\rm{\SVHN}}$ and $\dmdNegDataset=\validationSet_{\rm{\places}}$ for testing \SVHN and \places, respectively.
As such, \gls{dmd-a} is considered as an upper-bound, while \gls{dmd-u} is a realistic scenario where the distribution shift's nature cannot be anticipated.

Table \ref{tab: ood metrics} summarizes the results, showing that \gls{dmd-a} consistently yields the best \glspl{auc} for \places, but its performance greatly degrades for its \gls{dmd-u} version, demonstrating how the dataset used to fit the regressor impacts its detection capabilities. 
As originally observed in \cite{lee2018simple},  \gls{dmd}'s performance for convolutional models benefits from analyzing multiple layers (low performance on \gls{dmd-b}), while the improvement is marginal for transformer-based models, hardly justifying the required overhead.
Among the unsupervised methods, \gls{ourframework} and \gls{fs} consistently achieve the best performance, being comparable with each other. 

\begin{table}
  \centering
  \caption{\gls{ood} Detection performance in terms of \gls{auc} for \vgg{16} and \vit{B}{16}. \gls{id} samples correspond to \cifar{100}.}
  \label{tab: ood metrics}
  \begin{tabular}{lcccc}
    \toprule
    \multirow{2}{*}{\textbf{Method}} & \multicolumn{2}{c}{\textbf{\vgg{16}}} & \multicolumn{2}{c}{\textbf{\vit{B}{16}}} \\
    \cmidrule(lr){2-3} \cmidrule(lr){4-5}
    					& \places & \SVHN & \places & \SVHN  \\
    \midrule
    
    \gls{ourframework}	& 0.85 & 0.81 & 0.88 & 0.90 \\
    \gls{msp}  			& 0.84 & 0.82 & 0.84 & 0.90 \\
    \gls{doc}      		& 0.84 & 0.82 & 0.85 & 0.91 \\
    \gls{relu}       	& 0.74 & 0.64 & 0.67 & 0.67 \\
    \gls{fs}   			& 0.80 & 0.86 & 0.86 & 0.94 \\
    \gls{dmd-b}        	& 0.69 & 0.45 & 0.97 & 0.90 \\
    \gls{dmd-a}       	& 0.99 & 0.88 & 0.99 & 0.95 \\
    \gls{dmd-u}        	& 0.99 & 0.82 & 0.03 & 0.51 \\    
    \bottomrule
  \end{tabular}
\end{table}

\subsection{Attack Analysis}\label{sec: attack rejection}

For the final application, we evaluate how effectively \protoScore identifies \gls{aa} samples by reporting the corresponding \glspl{auc} in comparison to the reference methods.
Regarding sample selection, as in \cite{xu2017feature}, we only consider attacked samples for images correctly classified by the model, and that had their predicted label successfully changed by the attack.
For \gls{dmd-a}, the regressor is trained with samples from the attack used for testing, i.e., $\dmdNegDataset=\validationSet_{X}$ for testing each \gls{aa} $X\in\{\rm\gls{bim}, \gls{cw}, \gls{df}, \gls{pgd}\}$.
In turn \gls{dmd-u}'s regressor is trained using samples from all other \glspl{aa}, i.e., $\dmdNegDataset=\{\validationSet_{Y} ~\forall~ Y\in\{\rm\gls{bim}, \gls{cw}, \gls{df}, \gls{pgd}\} ~|~ Y\neq X\}$ for testing each \gls{aa} $X$, being a more realistic case since the \gls{aa} being tested is unknown.
%

%


The achieved results are reported in Table \ref{tab: attack_metrics}. 
\gls{cw} and \gls{df}, due to hyper-parameter choice, modify $\predictionProb$ such that the incorrect label attains only a slightly higher probability than the correct one, leading to low-confidence misclassification.
In contrast, \gls{bim} and \gls{pgd} push the model toward wrong predictions with high confidence.
It is straightforward to deduce that scores based solely on $\predictionProb(\img)$ (i.e., \gls{msp}, \gls{doc}, and \gls{relu}) will be effective at detecting \gls{cw} and \gls{df} as these attacks naturally produce low \gls{msp} values; however, their detection capability drops drastically under attacks like \gls{bim} and \gls{pgd}, which deliberately induce high-confidence errors, thereby compromising these confidence-based detectors by design.
As such, we argue that methods which analyze the intermediate activations (i.e., \gls{dmd} and \gls{ourframework}) are more robust to \glspl{aa} by design.
\gls{ourframework}'s score is low in both cases given that $\concept$ greatly mismatches the prediction's $\protoc$.
Since \gls{ourframework} takes into consideration also the last layer of the reference models, our results are influenced by it. However, the dependency over multiple layers allows \gls{ourframework} to be more robust to \gls{bim} and \gls{pgd} with respect to \gls{msp}, \gls{doc} and \gls{relu}.


\gls{dmd-a} and \gls{dmd-u} achieve high \glspl{auc} for \gls{df} and \gls{pgd}, and low scores for \gls{bim} and \gls{cw}.
Note that the pairs of attacks for which \gls{dmd} has high and low scores do not correspond to the separation of attacks based on their effects on the \gls{msp}, indicating that it is caused by the dataset dependency for training its regressor, reinforcing the idea that such a dependency can become a shortcoming of the method.


Different from the \gls{ood} case, \gls{dmd} shows significant improvements in its multi-layer version for both \vgg{16} and \vit{B}{16}, corroborating to the idea that methods which analyze the internal activations are inherently more robust to \glspl{aa}.

As expected, \gls{fs} outperforms other methods in this task, especially on \vgg{16}, while \gls{ourframework} is a valid alternative among the unsupervised methods. 
On \vit{B}{16}, \gls{ourframework} outperforms \gls{fs} on three attacks out of four.

 
\begin{table}
  \centering
  \caption{Attack Detection performance in terms of \gls{auc} under different attacks on \vgg{16} and \vit{B}{16}.}
  \begin{tabulary}{\linewidth}{LCCCCCCCC}
    \toprule
    \multirow{2}{*}{\textbf{Method}} & \multicolumn{4}{c}{\textbf{\vgg{16}}} & \multicolumn{4}{c}{\textbf{\vit{B}{16}}} \\
    \cmidrule(lr){2-5} \cmidrule(lr){6-9}
    					& \gls{bim} & \gls{cw} & \gls{df} & \gls{pgd} & \gls{bim} & \gls{cw} & \gls{df} & \gls{pgd} \\
    \midrule
    
    \gls{ourframework}	& 0.78 & 0.90 & 0.93 & 0.79 & 0.85 & 0.95 & 0.95 & 0.87 \\
    \gls{msp} 			& 0.51 & 0.96 & 0.96 & 0.50 & 0.56 & 0.95 & 0.93 & 0.55 \\
    \gls{doc}  			& 0.52 & 0.94 & 0.96 & 0.50 & 0.56 & 0.93 & 0.92 & 0.55 \\
    \gls{relu}			& 0.47 & 0.90 & 0.86 & 0.47 & 0.48 & 0.80 & 0.73 & 0.50 \\
    \gls{fs}   			& 0.89 & 0.97 & 0.89 & 0.91 & 0.83 & 0.97 & 0.91 & 0.81 \\
    \gls{dmd-b} 		& 0.42 & 0.47 & 0.57 & 0.56 & 0.75 & 0.74 & 0.87 & 0.80 \\
    \gls{dmd-a}			& 0.78 & 0.65 & 0.94 & 0.92 & 0.86 & 0.60 & 0.98 & 0.98 \\ 
    \gls{dmd-u} 		& 0.70 & 0.54 & 0.87 & 0.90 & 0.83 & 0.57 & 0.97 & 0.98 \\
    \bottomrule
  \end{tabulary}
  \label{tab: attack_metrics}
\end{table}

\subsection{Unknown Input Conditions}\label{subsec: AvA}

Up to this point, the different cases under investigation have been analyzed separately. 
However, in a practical scenario, it is impossible to know in advance whether inputs are \gls{id}, \gls{ood}, or \gls{aa}, making an approach robust for all three cases desirable.

To test this scenario, we compute a single threshold from \gls{id} samples as shown in Section \ref{subsec: confidence}, reflecting the impossibility of anticipating distribution shifts in \gls{ood} or the typology \gls{aa} samples.
Regarding \gls{dmd-a}, this evaluation exposes the shortcomings of being a supervised approach, requiring its regressor to be trained to distinguish correctly from incorrectly classified \gls{id} samples.


Accordingly, Tables \ref{tab:fpr_vgg16} and \ref{tab:fpr_vitb16} report the \fpr for \vgg{16} and \vit{B}{16}, respectively.  
Each column refers to a possible application and we highlight in bold the best-performing method and underline the second-best.
\gls{ourframework} consistently ranks among the best-performing methods across all scenarios, thereby providing the most favorable overall trade-off.

\begin{table*}
\centering
\caption{\fpr ($\downarrow$) given the threshold computed for \gls{id} analysis. 
$\text{mean}^\star$ is the mean of all {\fpr}s including the \gls{id} case, and \textit{mean} excludes the \gls{id} \fpr. 
\textbf{Bold} and \underline{underline} indicate the lowest and second lowest {\fpr}s, respectively.}
\label{tab:fpr_all}
\begin{subtable}{\textwidth}
\centering
\caption{\vgg{16}}
\label{tab:fpr_vgg16}
\begin{tabular}{lc|ccccc|cc|cccc|cc}
    \toprule
    & \multicolumn{1}{c}{} & \multicolumn{5}{c}{\textbf{Corruption}} & \multicolumn{2}{c}{\textbf{OOD}} & \multicolumn{4}{c}{\textbf{Attacks}} \\
    \cmidrule(lr){3-7} \cmidrule(lr){8-9} \cmidrule(lr){10-13}
    & ID & c0 & c1 & c2 & c3 & c4 & \SVHN & \places & \gls{bim} & \gls{cw} & \gls{df} & \gls{pgd} & mean & max\\ 
    \midrule
    \gls{ourframework} & \textbf{0.63} & \textbf{0.79} & \textbf{0.73} & \textbf{0.67} & \textbf{0.62} & \textbf{0.54} & \textbf{0.46} & \textbf{0.44} & \underline{0.77} & 0.50 & 0.36 & \underline{0.73} & \textbf{0.60} & \textbf{0.79}\\
    \gls{msp}          & 0.65 & \underline{0.81} & 0.76 & 0.74 & \underline{0.71} & 0.68 & 0.57 & 0.50 & 0.82 & \underline{0.30} & \textbf{0.19} & 0.85 & \underline{0.63} & 0.85\\
    \gls{doc}          & \underline{0.64} & \textbf{0.79} & \underline{0.74} & \underline{0.71} & 0.68 & \underline{0.65} & 0.53 & \underline{0.45} & 0.81 & 0.58 & \underline{0.22} & 0.84 & \underline{0.63} & \underline{0.84}\\
    \gls{relu}         & 0.80 & 0.90 & 0.90 & 0.90 & 0.91 & 0.91 & 0.98 & 0.90 & 0.98 & 0.68 & 0.85 & 0.99 & 0.89 & 0.99\\
    \gls{fs}           & 0.76  & 0.84 & 0.79 & 0.77 & 0.74 & 0.70 & \underline{0.47} & 0.67 & \textbf{0.28} & \textbf{0.16} & 0.67 & \textbf{0.32} & \textbf{0.60} & 0.84\\
    \gls{dmd-b}        & 0.97 & 0.97 & 0.97 & 0.98 & 0.97 & 0.98 & 1.00 & 0.89 & 0.98 & 0.96 & 0.96 & 0.96 & 0.96 & 0.98\\
    \gls{dmd-a}          & 0.92 & 0.90 & 0.82 & 0.76 & 0.73 & 0.67 & 0.75 & 0.83 & 0.95 & 0.95 & 0.84 & 0.85 & 0.83 & 0.95\\
    \bottomrule
\end{tabular}
\end{subtable}

\vspace{0.5cm}

\begin{subtable}{\textwidth}
\centering
\caption{\vit{B}{16}}
\label{tab:fpr_vitb16}
\begin{tabular}{lc|ccccc|cc|cccc|cc}
    \toprule
    & \multicolumn{1}{c}{} & \multicolumn{5}{c}{\textbf{Corruption}} & \multicolumn{2}{c}{\textbf{OOD}} & \multicolumn{4}{c}{\textbf{Attacks}} \\
    \cmidrule(lr){3-7} \cmidrule(lr){8-9} \cmidrule(lr){10-13}
    & ID & c0 & c1 & c2 & c3 & c4 & \SVHN & \places & \gls{bim} & \gls{cw} & \gls{df} & \gls{pgd} & mean & max\\ 
    \midrule
    \gls{ourframework} & \textbf{0.55} & \textbf{0.81} & \textbf{0.75} & \textbf{0.71} & \textbf{0.66} & \textbf{0.57} & 0.34 & \underline{0.40} & \underline{0.51} & \underline{0.30} & \textbf{0.23} & \underline{0.53} & \textbf{0.53} & \textbf{0.81}\\
    \gls{msp}          & \underline{0.57} & \underline{0.84} & 0.79 & 0.75 & 0.71 & 0.65 & 0.34 & 0.49 & 0.69 & \underline{0.30} & \underline{0.30} & 0.71 & 0.60 & \underline{0.84}\\
    \gls{doc}          & 0.58 & \underline{0.84} & \underline{0.78} & \underline{0.74} & \underline{0.70} & \underline{0.63} & \underline{0.28} & 0.45 & 0.67 & 0.60 & 0.39 & 0.70 & 0.62 & \underline{0.84}\\
    \gls{relu}         & 0.78 & 0.92 & 0.92 & 0.92 & 0.92 & 0.93 & 1.00 & 0.95 & 0.98 & 0.85 & 0.92 & 0.98 & 0.94 & 0.98\\
    \gls{fs}           & 0.75 & 0.88 & 0.85 & 0.82 & 0.80 & 0.73 & \textbf{0.20} & 0.51 & \textbf{0.37} & \textbf{0.16} & 0.43 & \textbf{0.48} & \underline{0.57} & 0.88\\
    \gls{dmd-b}        & 0.83 & 0.90 & 0.88 & 0.86 & 0.83 & 0.78 & 0.55 & \textbf{0.11} & 0.64 & 0.82 & 0.54 & 0.61 & 0.68 & 0.90\\
    \gls{dmd-a}          & 0.90 & 0.94 & 0.93 & 0.92 & 0.91 & 0.89 & 0.70 & 0.96 & 0.95 & 0.94 & 0.93 & 0.92 & 0.91 & 0.96\\
    \bottomrule
\end{tabular}
\end{subtable}
\end{table*}

\subsection{Computation Complexity and Time}\label{subsec: complexity}

To summarize this aspect, Table \ref{tab: complexity} shows the average computation time overhead rescaled by the average inference time for each method, considering only their online steps on an NVIDIA H100 device with $80$GB of RAM.
Regarding \gls{ourframework}'s inference overhead, the dimensionality reduction step, a.k.a, computing eq. \eqref{eq: corevector}, corresponds to roughly $90\%$ and $60\%$ of the overhead for \vgg{16} and \vit{B}{16}, respectively, which is caused by the higher activation sizes of convolutional layers.

Another advantage of \gls{ourframework} is its low online computational cost. All expensive operations, namely the layers’ parameter \gls{svd} decomposition ($\SVDVtrim$), clustering of the training-set corevectors ($\coreVec$), and the empirical association between \gls{llf}, \gls{hlf}, and the \protoclasses ($\empiricalPosterior$, $\protoc$), are performed once offline. At inference time, \gls{ourframework} requires only lightweight matrix multiplications and a cosine similarity over compact representations.

In contrast, \gls{dmd} requires a back-propagation step for each evaluated sample, leading to a substantially higher and model-dependent computational burden.
This difference is especially evident in \gls{dmd-b}, whose best configuration according to our tests uses backpropagation for \vgg{16}, but does not use it for \vit{B}{16}.
Note that \gls{dmd-a} and \gls{dmd-u} differ only in the datasets used for training their regressors, so their computational overhead is the same.

\gls{doc} and \gls{relu}, due to hyperparameter settings incur minimal overhead. \gls{fs} remains the most expensive method due to its \gls{nlm} filtering.

Table \ref{tab: complexity} summarizes the average online overhead of each method, normalized by their respective inference times on an NVIDIA H100 ($80$ GB RAM). For \gls{ourframework}, the dimensionality-reduction step (eq. \eqref{eq: corevector}) dominates the cost, accounting for approximately  $90\%$ of the overhead on \vgg{16} and $60\%$ on \vit{B}{16}, reflecting the larger activation sizes of convolutional layers.

\begin{table}
	\centering
	\caption{Average computation time overhead per sample normalized according to the average model inference time.}
	\label{tab: complexity}
	\begin{tabulary}{\linewidth}{lcc}
		\toprule
		\textbf{Method} & \textbf{\vgg{16}} & \textbf{\vit{B}{16}} \\
		\midrule
		
		\gls{ourframework}	& $2.82\pm0.6$ & $(9.5\pm0.3)\times10^{-1}$ \\
		\gls{doc} 			& $(1.8\pm0.1)\times10^{-4}$ & $(1.1\pm0.8)\times10^{-4}$ \\
		\gls{relu}			& $(4.1\pm0.3)\times10^{-3}$ & $(1.6\pm0.2)\times10^{-3}$ \\
		\gls{fs}   			& $(2.9\pm0.3)\times10^{2}$ & $(1.8\pm0.6)\times10^{2}$ \\
		\gls{dmd}			& $(2.5\pm0.2)\times10^{1}$ & $(4.7\pm0.6)\times10^{1}$ \\
        \gls{dmd-b}			& $1.2\pm0.3$ & $(6.1\pm0.3)\times10^{-2}$ \\
		\bottomrule
	\end{tabulary}
\end{table}

	\section{Conclusions} \label{sec: conclusions} 

This work introduced \gls{ourframework}, a post-hoc method for estimating prediction confidence in generic pre-trained \glspl{dnn} by analyzing their internal activations. The resulting score enables confidence estimation, \gls{ood} detection, and \gls{aa} detection within a unified and lightweight framework. \gls{ourframework} incorporates dimensionality reduction for scalability, unsupervised clustering to identify activation-level features, and an empirical association step linking these features to semantic labels.

We evaluated the method on \vgg{16}, \vit{B}{16}, and the \cifar{100}, \cifar{100C}, \SVHN, and \places datasets, comparing it against \gls{doc}, \gls{relu}, and the intermediate-activation method \gls{dmd}. Beyond matching or outperforming these baselines in individual tasks, \gls{ourframework} provides consistently reliable behavior across all scenarios without requiring task-specific supervision.

The results reveal a key distinction: while output-based approaches excel when errors coincide with low softmax confidence, they fail under high-confidence \glspl{aa} or \gls{ood} shifts. In contrast, activation-based methods capture deviations in the internal decision process, enabling more robust detection even when the output layer is misleading. \gls{ourframework} exploits this advantage while avoiding the supervised regressor required by prior techniques.

Future directions include extending the framework to non-vision domains, exploring adaptive clustering strategies, and integrating \gls{ourframework} into training-time feedback loops to enhance calibration and robustness.

	\bibliographystyle{IEEEtran}
	\bibliography{bib/IEEEabrv,bib/XAI}

    \vskip -2\baselineskip plus -2fil

\begin{IEEEbiography}[{\includegraphics[width=1in,height=1.25in,clip,keepaspectratio]{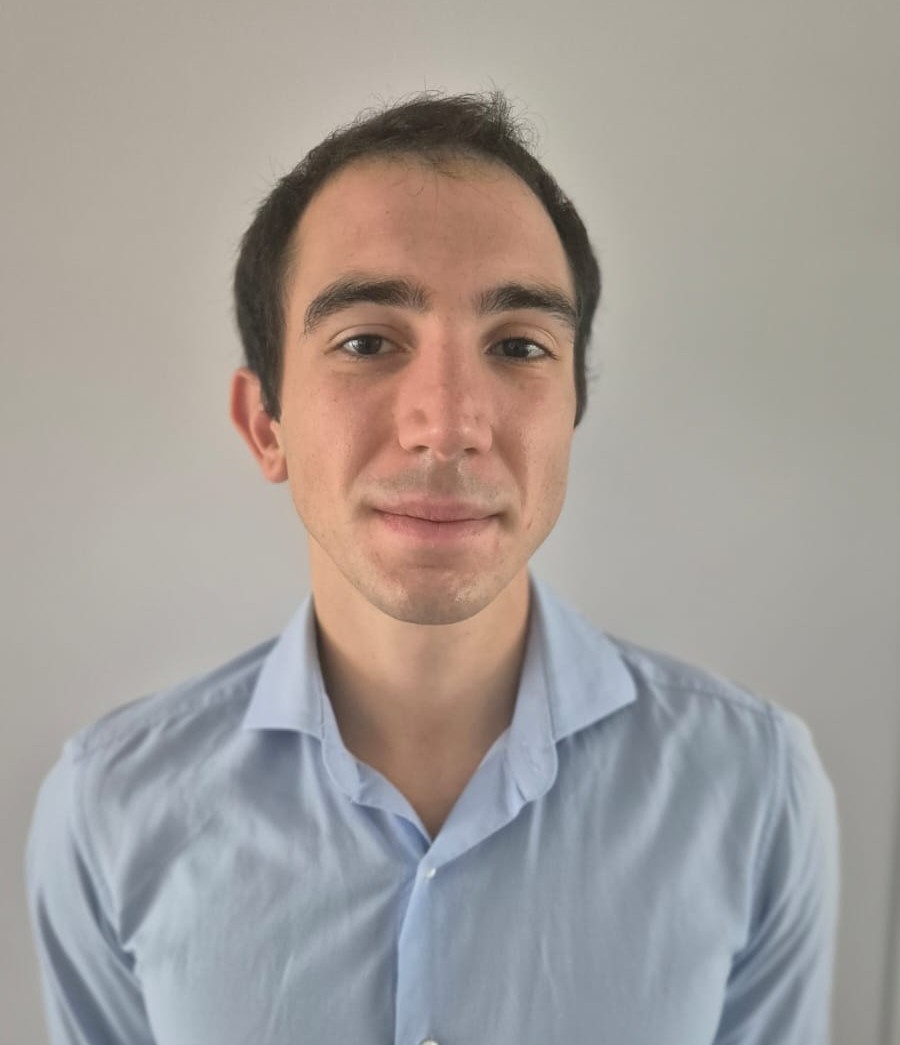}}]{Lorenzo Capelli}
	He received his B.Sc. (with Honors) in biomedical engineering in 2021 and his M.Sc. (with Honors) in electronic engineering from the University of Bologna in 2024. He is currently pursuing a Ph.D. in Engineering and Information Technology for Structural and Environmental Monitoring and Risk Management within the Statistical Signal Processing Group. His research focuses on signal processing, explainable AI, and model trustworthiness.
\end{IEEEbiography}

\vskip -2\baselineskip plus -2fil

\begin{IEEEbiography}[{\includegraphics[width=1in,height=1.25in,clip,keepaspectratio]{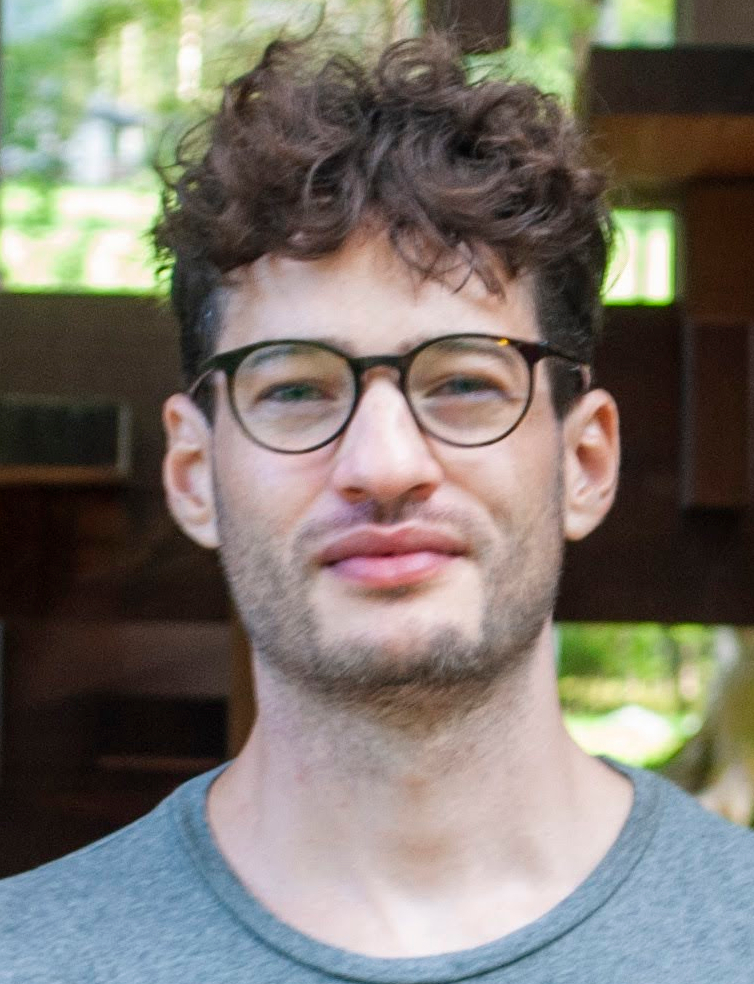}}]{Leandro de Souza Rosa}
	has a BSc (2013) in Computer Engineering and obtained a Ph.D. in 2019 from the Institute of Mathematics and Computer Sciences at The University of São Paulo, Brazil, which was partially developed at the Department of Electrical and Electronic Engineering, Imperial College London, U.K. He worked as a postdoc researcher at the Istituto Italiano di Tecnologia (Genova, Italy), in the Faculty of Mechanical, Maritime and Materials Engineering, Delft University of Technology (Delft, Netherlands), and La Sapienza University of Rome (Rome, Italy). He currently holds a researcher position in the Department of Electrical, Electronic and Information Engineering of the University of Bologna (Bologna, Italy).
    His research interests include hardware design, perception, robotics, and machine learning, focusing on interactions between humans and autonomous systems.
\end{IEEEbiography}

\vfill

\begin{IEEEbiography}[{\includegraphics[width=1in, height=1.25in, clip, keepaspectratio]{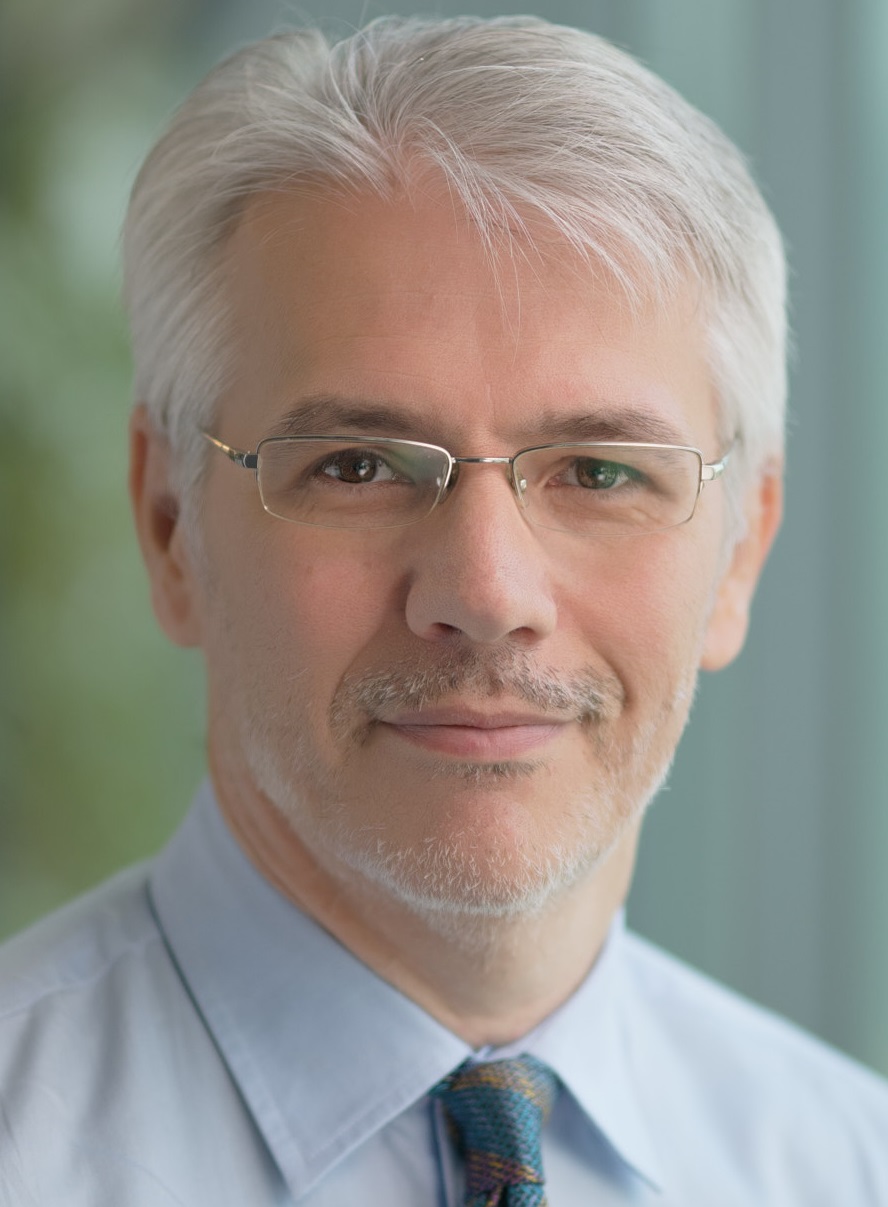}}]{Gianluca Setti} is the Dean of the Computer, Electrical, Mathematical Sciences and Engineering Division and a Professor of Electrical and Computer Engineering at KAUST. Before moving to KAUST, he was with the University of Ferrara (1997-2017) and with Politecnico di Torino (2017-2022) as a Professor of Electronics, Signal and Data Processing. He held also several positions as Visiting Professor/Scientist at EPFL (2002, 2005), UCSD (2004), IBM (2004, 2007) and at University of Washington (2008, 2010).
He served as the Editor-in-Chief for the IEEE Transactions on Circuits and Systems - Part II (2006-2007) and of the IEEE Transactions on Circuits and Systems - Part I (2008-2009). Since 2019 he is the first non-US Editor-in-Chief of the Proceedings of the IEEE, the flagship journal of the IEEE.
In 2010, he served as IEEE CAS Society President. In 2013-2014 he was the first non North-American Vice President of the IEEE for Publication Services and Products.
Dr. Setti received several awards, including the 2004 IEEE CAS Society Darlington Award, 2013 IEEE CAS Society Meritorious Service Award, 2013 IEEE CAS Society Guillemin-Cauer Award, and the 2019 IEEE Transactions on Circuits and Systems Best Paper Award. Since 2016, he is also a Fellow of the IEEE.
His research interests include recurrent neural networks, EMC, compressive sensing and statistical signal processing, biomedical circuits and systems, power electronics, IoT, circuits and systems for machine learning, and applications of AI techniques for anomaly detection and predictive maintenance.
\end{IEEEbiography}

\vskip -2\baselineskip plus -2fil

\begin{IEEEbiography}[{\includegraphics[width=1in,height=1.25in,clip,keepaspectratio]{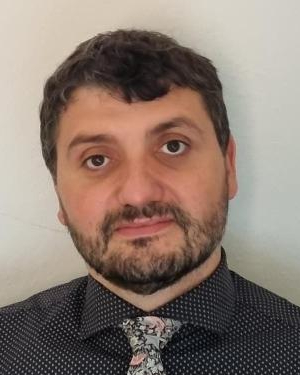}}]{Mauro Mangia}
received the B.Sc. and M.Sc. degrees in electronic engineering and the Ph.D. degree in information technology from the University of Bologna, Bologna, Italy, in 2005, 2009, and 2013, respectively. He was a Visiting Ph.D. Student at the École Polytechnique Fédérale de Lausanne in 2009 and 2012. He is currently an Associate Professor with the Department of Electrical, Electronic and Information Engineering of the University of Bologna within the Statistical Signal Processing Group. He is also a member of both the Advanced Research Center for Electronic Systems (ARCES) and the Alma Mater Research Institute for Human-centered Artificial Intelligence.
He is the scientific coordinator of several international research projects funded by public agencies, including the Italian Ministry of University and Research, the Italian Space Agency (ASI), and the European Space Agency (ESA), as well as projects supported by private entities such as Rete Ferroviaria Italiana (RFI) and Thales Alenia Space. His research interests include nonlinear systems, explainable artificial intelligence, machine learning and AI, anomaly detection, Internet of Things, Big Data analytics, and optimization.
He was a recipient of the 2013 IEEE CAS Society Guillemin–Cauer Award and of the 2019 IEEE BioCAS Transactions Best Paper Award. He received the Best Student Paper Award at the IEEE International Symposium on Circuits and Systems (ISCAS 2011, as student) and at the IEEE International Instrumentation and Measurement Technology Conference (I2MTC 2024, as supervisor).
\end{IEEEbiography}

\vskip -2\baselineskip plus -2fil

\begin{IEEEbiography}[{\includegraphics[width=1in, height=1.25in, clip, keepaspectratio]{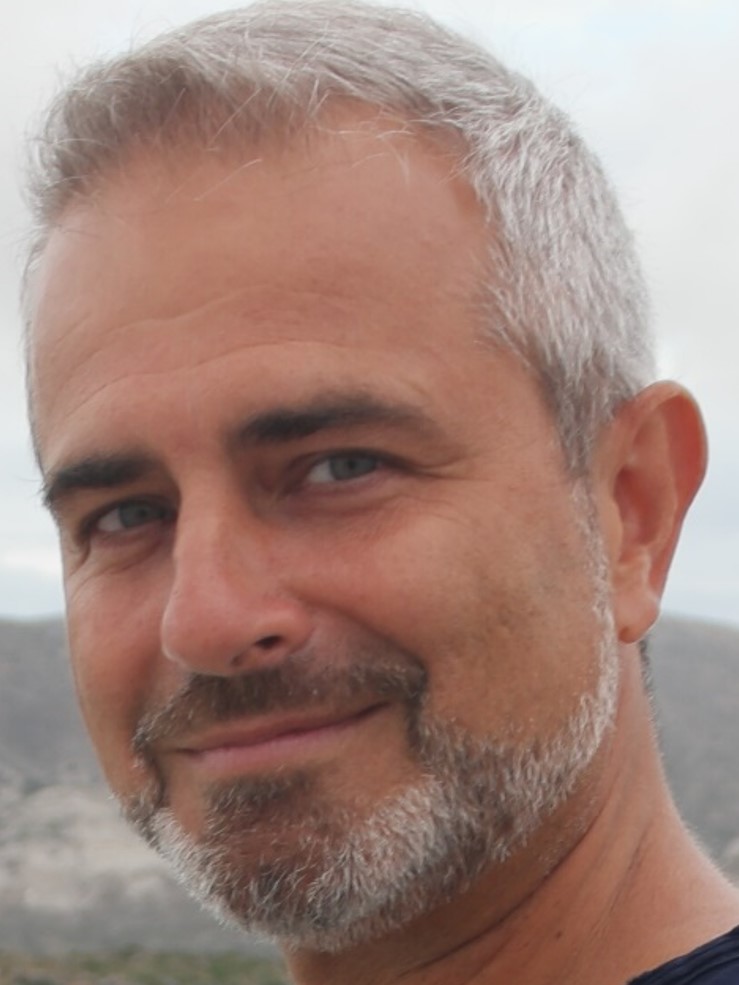}}]{Riccardo Rovatti} (M'99-SM'02-F'12) received the M.S. degree in electronic engineering and the Ph.D. degree in electronics, computer science, and telecommunications from the University of Bologna, Italy, in 1992 and 1996, respectively. He is currently a Full Professor of electronics at the University of Bologna. He has authored more than 300 technical contributions to international conferences and journals and two volumes. His research focuses on mathematical and applicative aspects of statistical signal processing, on machine learning for signal processing, and on the application of statistics to nonlinear dynamical systems.
He was a Distinguished Lecturer of the IEEE CAS Society for the years 2017-2018. He was a recipient of the 2004 IEEE CAS Society Darlington Award, the 2013 IEEE CAS Society Guillemin-Cauer Award, and the 2019 IEEE BioCAS Transactions Best Paper Award. He received the Best Paper Award at ECCTD 2005 and the Best Student Paper Award at the EMC Zurich 2005 and at the International Symposium On Circuits and Systems (ISCAS) 2011. He is an IEEE fellow for his contribution to nonlinear and statistical signal processing applied to electronic systems.

\vspace*{\fill}%

\end{IEEEbiography}    
\end{document}